\documentclass[journal]{IEEEtran}

\usepackage{graphicx}
\usepackage{amsmath}
\usepackage{url}
\usepackage{cite}

\renewcommand{\baselinestretch}{1.0}

\usepackage{type1cm}

\usepackage{algorithm}
\usepackage{algpseudocode}
\algrenewcommand{\algorithmicreturn}{\State \textbf{return}}
\renewcommand{\algorithmiccomment}[1]{// #1}
\begin{document}


\title{Spatial Concept Acquisition for a Mobile Robot that Integrates Self-Localization and Unsupervised Word Discovery from Spoken Sentences}

%
%
%
\author{Akira Taniguchi, 
        Tadahiro Taniguchi,~\IEEEmembership{Member,~IEEE,} 
        and Tetsunari Inamura,~\IEEEmembership{Member,~IEEE,}%

\thanks{Akira Taniguchi and Tadahiro Taniguchi are with Ritsumeikan University, 1-1-1 Noji Higashi, Kusatsu, Shiga 525-8577, Japan (e-mail:{a.taniguchi@em.ci.ritsumei.ac.jp}; {taniguchi@em.ci.ritsumei.ac.jp}).}
\thanks{Tetsunari Inamura is with National Institute of Informatics/The Graduate University for Advanced Studies,
        2-1-2 Hitotsubashi, Chiyoda-ku, Tokyo 101-8430, Japan 
        (e-mail:{inamura@nii.ac.jp}).}
}

\maketitle



\begin{abstract}
In this paper, we propose a novel unsupervised learning method for the lexical acquisition of words related to places visited by robots, from human continuous speech signals. 
We address the problem of learning novel words by a robot that has no prior knowledge of these words except for a primitive acoustic model.
Further, we propose a method that allows a robot to effectively use the learned words and their meanings for self-localization tasks.
The proposed method is nonparametric Bayesian spatial concept acquisition method (SpCoA) that integrates the generative model for self-localization and the unsupervised word segmentation in uttered sentences via latent variables related to the spatial concept.
We implemented the proposed method SpCoA on SIGVerse, which is a simulation environment, and TurtleBot\,2, which is a mobile robot in a real environment. 
Further, we conducted experiments for evaluating the performance of SpCoA.
The experimental results showed that SpCoA enabled the robot to acquire the names of places from speech sentences.
They also revealed that the robot could effectively utilize the acquired spatial concepts and reduce the uncertainty in self-localization. 
\end{abstract}


%
\IEEEpeerreviewmaketitle

\section{Introduction}
\label{sec:introduction} 
\IEEEPARstart{A}{utonomous} robots, such as service robots, operating in the human living environment with humans have to be able to perform various tasks and language communication. 
To this end, robots are required to acquire novel concepts and vocabulary on the basis of the information obtained from their sensors, e.g., laser sensors, microphones, and cameras, and recognize a variety of objects, places, and situations in an ambient environment. 
Above all, we consider it important for the robot to learn the names that humans associate with places in the environment and the spatial areas corresponding to these names; i.e., the robot has to be able to understand words related to places. 
Therefore, it is important to deal with considerable uncertainty, such as the robot's movement errors, sensor noise, and speech recognition errors.

Several studies on language acquisition by robots have assumed that robots have no prior lexical knowledge.
These studies differ from speech recognition studies based on a large vocabulary and natural language processing studies based on lexical, syntactic, and semantic knowledge \cite{roy2002learning,taguchi2009learning}.
Studies on language acquisition by robots also constitute a constructive approach to the human developmental process and the emergence of symbols.

\begin{figure}[!tb]
  \begin{center}
    \includegraphics[width=250pt]{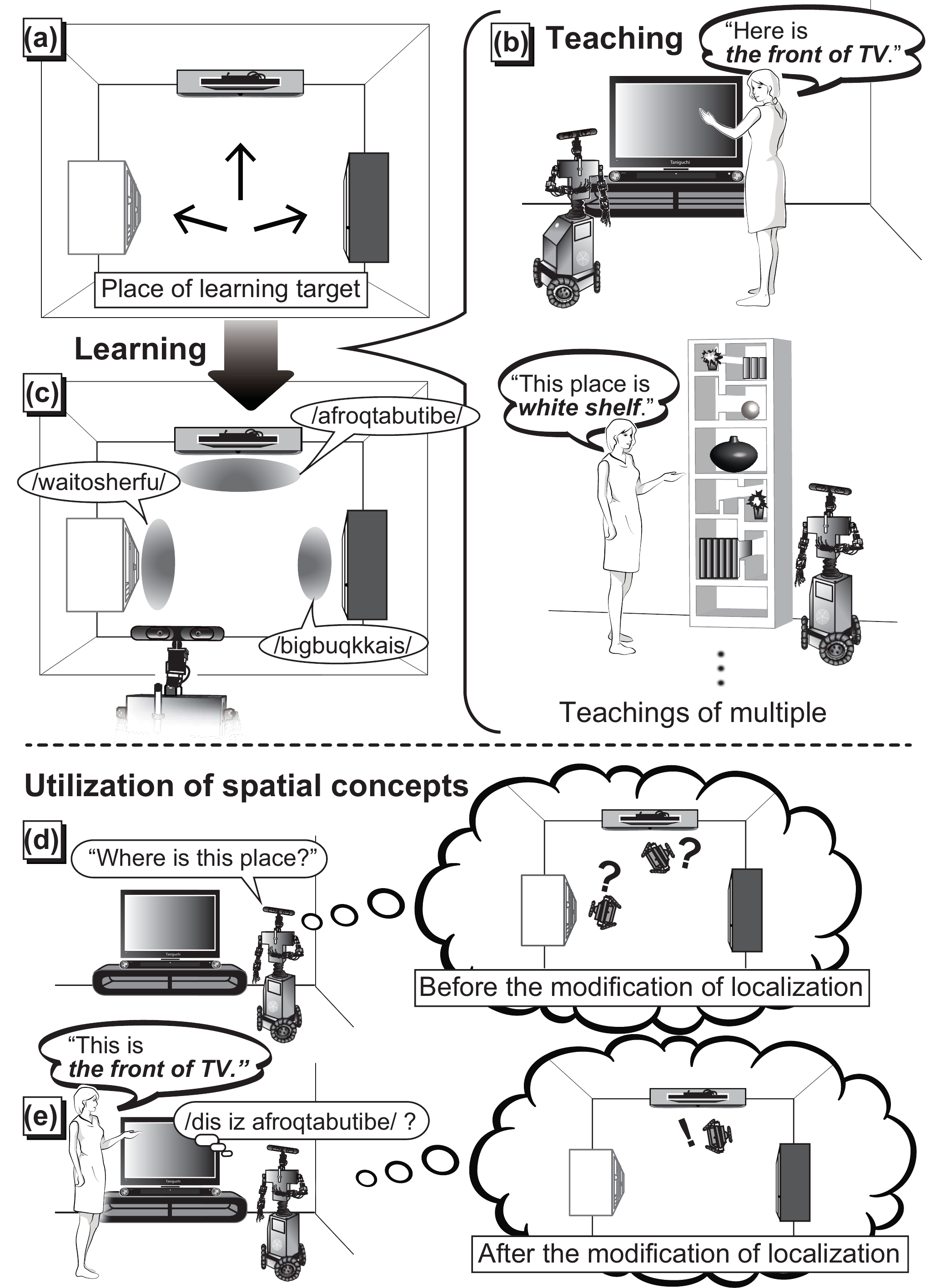}
    \caption{Schematic representation of the target task: 
(a)\,Learning targets are three places near the objects. 
(b)\,When an utterer and a robot came in front of the TV, the utterer spoke {\it ``Here is the front of TV.''} 
The same holds true for the other places. 
(c)\,The robot performs word discovery from the uttered sentences and learns the related spatial concepts. 
Words are related to each place. 
(d)\,
The robot is in front of TV actually.
However, the hypothesis of the self-position of the robot is uncertain. 
Then, the robot asks a neighbor what the current place is. 
(e)\,
By utilizing spatial concepts and an uttered sentence, the robot can narrow down the hypothesis of self-position.}
    \label{fig:teian}
  \end{center}
\end{figure}

The objectives of this study were to build a robot that learns words related to places and efficiently utilizes this learned vocabulary in self-localization.
Lexical acquisition related to places is expected to enable a robot to improve its spatial cognition.
A schematic representation depicting the target task of this study 
is shown in Fig.~\ref{fig:teian}.
This study assumes that a robot does not have any vocabularies in advance but can recognize syllables or phonemes. 
The robot then performs self-localization while moving around in the environment, as shown in Fig.~\ref{fig:teian} (a).
An utterer speaks a sentence including the name of the place to the robot, as shown in Fig.~\ref{fig:teian} (b).
For the purposes of this study, we need to consider the problems of self-localization and lexical acquisition simultaneously.

When a robot learns novel words from utterances, it is difficult to determine segmentation boundaries and the identity of different phoneme sequences from the speech recognition results, which can lead to errors.
First, let us consider the case of the lexical acquisition of an isolated word.
For example, if a robot obtains the speech recognition results {\it ``aporu''}, {\it ``epou''}, and {\it ``aqpuru''} (incorrect phoneme recognition of {\it apple}), it is difficult for the robot to determine whether they denote the same referent without prior knowledge. 
Second, let us consider a case of the lexical acquisition of the utterance of a sentence.
For example, a robot obtains a speech recognition result, such as {\it ``thisizanaporu.''}
The robot has to necessarily segment a sentence into individual words, e.g., {\it ``this''}, {\it ``iz''}, {\it ``an''}, and {\it ``aporu''}. 
In addition, it is necessary for the robot to recognize words referring to the same referent, e.g., the fruit {\it apple}, from among the many segmented results that contain errors.
In case of Fig.~\ref{fig:teian} (c), there is some possibility of learning names including phoneme errors, e.g., {\it ``afroqtabutibe,''} because the robot does not have any lexical knowledge. 

On the other hand, when a robot performs online probabilistic self-localization, we assume that the robot uses sensor data and control data, e.g., values obtained using a range sensor and odometry.
If the position of the robot on the global map is unclear, the difficulties associated with the identification of the self-position by only using local sensor information become problematic.
In the case of global localization using local information, e.g., a range sensor, the problem that the hypothesis of self-position is present in multiple remote locations, frequently occurs, as shown in Fig.~\ref{fig:teian} (d).

In order to solve the abovementioned problems, in this study, we adopted the following approach. 
An utterance is recognized as not a single phoneme sequence but a set of candidates of multiple phonemes.
We attempt to suppress the variability in the speech recognition results by performing word discovery taking into account the multiple candidates of speech recognition.
In addition, the names of places are learned by associating with words and positions.
The lexical acquisition is complemented by using certain particular spatial information; i.e., this information is obtained by hearing utterances including the same word in the same place many times.
Furthermore, in this study, we attempt to address the problem of the uncertainty of self-localization by improving the self-position errors by using a recognized utterance including the name of the current place and the acquired spatial concepts, as shown in Fig.~\ref{fig:teian} (e).


In this paper, 
we propose {\it nonparametric Bayesian spatial concept acquisition method} (SpCoA) on basis of unsupervised word segmentation and a nonparametric Bayesian generative model that integrates self-localization and a clustering in both words and places.
The main contributions of this paper are as follows:
\begin{itemize}
\item 
We have proposed a learning method for spatial concepts that can perform the lexical acquisition related to places, i.e., the names of places, from a continuous speech signal in an unsupervised manner.
\item 
We have achieved relatively accurate lexical acquisition that reduced the variability and errors in phonemes by performing word discovery using the multiple candidates of the speech recognition results, i.e., by using a lattice format.
\item 
In addition to the general self-localization method of mobile robots, we showed that self-localization by the proposed method can reduce the uncertainty of self-position by utilizing the learned spatial concepts and an uttered sentence about the current position.
\end{itemize}

The remainder of this paper is organized as follows:
In Section~\ref{sec:senkou}, previous studies on language acquisition and lexical acquisition relevant to our study are described.
In Section~\ref{sec:model}, the proposed method SpCoA is presented.
In Sections~\ref{sec:experiments1} and \ref{sec:experiments3}, we discuss the effectiveness of SpCoA in the simulation and in the real environment.
Section~\ref{sec:conclusion} concludes this paper.


\section{Related Works}
\label{sec:senkou}
\subsection{Lexical acquisition}
\label{sec:senkou:goi}
Most studies on lexical acquisition typically focus on lexicons about objects \cite{roy2002learning,iwahashi2003language,Iwahashi2007,iwahashi2009robots,gorniak2005probabilistic,qu2008incorporating,qu2010context,hornstein2010multimodal,nakamura2011,araki2012online}.
Many of these studies have not be able to address the lexical acquisition of words other than those related to objects, e.g., words about places. 

Roy et al. proposed a computational model that enables a robot to learn the names of objects from an object image and spontaneous infant-directed speech \cite {roy2002learning}. 
Their results showed that the model performed speech segmentation, word discovery, and visual categorization. 
Iwahashi et al. reported that a robot properly understands the situation and acquires the relationship of object behaviors and sentences \cite{iwahashi2003language,Iwahashi2007,iwahashi2009robots}.
Qu {\&} Chai focused on the conjunction between speech and eye gaze and the use of domain knowledge in lexical acquisition {\cite{qu2008incorporating,qu2010context}}. 
They proposed an unsupervised learning method that automatically acquires novel words for an interactive system.
Qu {\&} Chai's method based on the IBM translation model {\cite{brown1993mathematics}} estimates the word-entity association probability.

Nakamura et al. proposed a method to learn object concepts and word meanings from multimodal information and verbal information \cite{nakamura2011}. 
The method proposed in \cite{nakamura2011} is a categorization method based on multimodal latent Dirichlet allocation (MLDA) that enables the acquisition of object concepts from multimodal information, such as visual, auditory, and haptic information \cite{nakamura2011multimodal}.
Araki et al. addressed the development of a method combining unsupervised word segmentation from uttered sentences by a nested Pitman-Yor language model (NPYLM) \cite{mochihashi2009bayesian} and the learning of object concepts by MLDA \cite{araki2012online}.
However, the disadvantage of using NPYLM was that phoneme sequences with errors did not result in appropriate word segmentation.


These studies did not address the lexical acquisition of the space and place that can also tolerate the uncertainty of phoneme recognition.
However, for the introduction of robots into the human living environment, 
robots need to acquire a lexicon related to not only objects but also places.
Our study focuses on the lexical acquisition related to places.
Robots can adaptively learn the names of places in various human living environments by using SpCoA.
We consider that the acquired names of places can be useful for various tasks, e.g., tasks with a movement of robots by the speech instruction.
%

\subsection{Simultaneous learning of places and vocabulary}
\label{sec:senkou:douji}
The following studies have addressed lexical acquisition related to places.
However, these studies could not utilize the learned language knowledge in other estimations such as the self-localization of a robot.

Taguchi et al. proposed a method for the unsupervised learning of phoneme sequences and relationships between words and objects from various user utterances without any prior linguistic knowledge other than an acoustic model of phonemes \cite{taguchi2009learning,taguchi2012learning}.
Further, they proposed a method for the simultaneous categorization of self-position coordinates and lexical learning \cite{taguchi2011learning}. 
These experimental results showed that it was possible to learn the name of a place from utterances in some cases and to output words corresponding to places in a location that was not used for learning.

Milford et al. proposed RatSLAM inspired by the biological knowledge of a pose cell of the hippocampus of rodents \cite{milford2004ratslam}.
Milford et al. proposed a method that enables a robot to acquire spatial concepts by using RatSLAM \cite{milford2007learning}. 
Further, Lingodroids, mobile robots that learn a language through robot-to-robot communication, have been studied \cite{schulz2011lingodroids,heath2013communication,schulz2011we}.
Here, a robot communicated the name of a place to other robots at various locations.
Experimental results showed that two robots acquired the lexicon of places that they had in common.
In \cite{schulz2011we}, the researchers showed that it was possible to learn temporal concepts in a manner analogous to the acquisition of spatial concepts.
These studies reported that the robots created their own vocabulary.
However, these studies did not consider the acquisition of a lexicon by human-to-robot speech interactions.

Welke et al. proposed a method that acquires spatial representation by the integration of the representation of the continuous state space on the sensorimotor level and the discrete symbolic entities used in high-level reasoning \cite{welke2013grounded}.
This method estimates the probable spatial domain and word from the given objects by using the spatial lexical knowledge extracted from Google Corpus and the position information of the object.
Their study is different from ours because their study did not consider lexicon learning from human speech.

In the case of global localization, the hypothesis of self-position often remains in multiple remote places.
In this case, there is some possibility of performing an incorrect estimation and increasing the estimation error.
This problem exists during teaching tasks and self-localization after the lexical acquisition.
The abovementioned studies could not deal with this problem.
In this paper, we have proposed a method that enables a robot to perform more accurate self-localization by reducing the estimation error of the teaching time by using a smoothing method in the teaching task and by utilizing words acquired through the lexical acquisition.
The strengths of this study are that learning of spatial concept and self-localization represented as one generative model and robots are able to utilize acquired lexicon to self-localization autonomously.


\section{Spatial Concept Acquisition}
\label{sec:model}

We propose {\it nonparametric Bayesian spatial concept acquisition method} (SpCoA) that integrates a nonparametric morphological analyzer for the lattice \cite{neubig2012bayesian}, i.e., latticelm\footnote{latticelm is the name of the tool that \cite{neubig2012bayesian} is implemented and is treated as the name of the method in this study.
 \url{http://www.phontron.com/latticelm/}}, a spatial clustering method, and Monte Carlo localization (MCL) \cite{dellaert1999monte}.

\subsection{Generative model}

In our study, we define a {\it position} as a specific coordinate or a local point in the environment, and the {\it position distribution} as the spatial area of the environment.
Further, we define a {\it spatial concept} as the names of places and the position distributions corresponding to these names.

\begin{figure}[tb]
  \begin{center}
    \includegraphics[width=170pt]{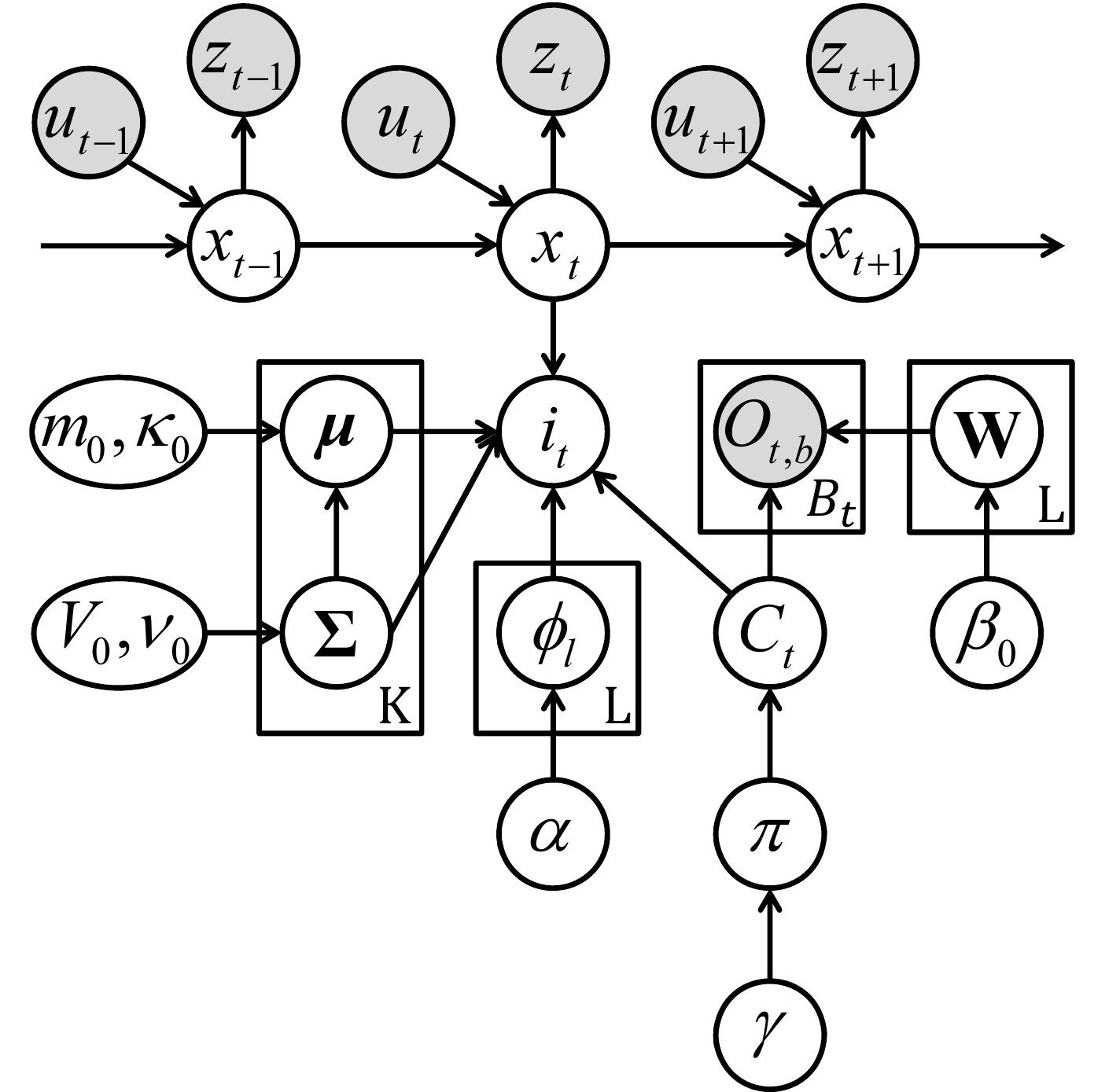}
    \caption{Graphical model of the proposed method SpCoA}
    \label{fig:graphicalmodel}
  \end{center}
\end{figure}

The model that was developed for spatial concept acquisition is a probabilistic generative model that integrates a self-localization with the simultaneous clustering of places and words.
Fig.~\ref{fig:graphicalmodel} shows the graphical model for spatial concept acquisition.
Table~\ref{youso} shows each variable of the graphical model.
The number of words in a sentence at time $t$ is denoted as $B_{t}$.
The generative model of the proposed method is defined as equation (\ref{eq:seisei1}-\ref{eq:seisei10}).

\begin{eqnarray}
\pi &\sim& {\rm GEM}(\gamma ) \label{eq:seisei1} \\
C_{t} &\sim& {\rm Mult}(\pi) \label{eq:seisei2} \\
W &\sim& {\rm Dir}(\beta _{0}) \label{eq:seisei3} \\
O_{t,b} &\sim& {\rm Mult}(W_{C_{t}}) \label{eq:seisei4} \\
\phi _{l} &\sim& {\rm GEM}(\alpha ) \label{eq:seisei5} \\
i_{t} &\sim& p(i_{t} \mid x_{t},\mbox{\boldmath{$\mu $}},\mathbf{\Sigma} ,\phi _{l},C_{t}) \label{eq:seisei6} \\
\Sigma &\sim& {\cal IW}( \Sigma \mid V_{0} ,\nu _{0} )  \label{eq:seisei7} \\
\mu &\sim& {\cal N}( \mu  \mid m_{0}, ( \Sigma / \kappa _{0} )) \label{eq:seisei8} \\
x_{t} &\sim& p(x_{t} \mid x_{t-1},u_{t}) \label{eq:seisei9} \\
z_{t} &\sim& p(z_{t} \mid x_{t}) \label{eq:seisei10}
\label{eq:seisei}
\end{eqnarray}
Then, the probability distribution for equation (\ref{eq:seisei6}) can be defined as follows:
\begin{eqnarray}
\lefteqn{
p(i_{t} \mid x_{t},\mbox{\boldmath{$\mu $}},\mathbf{\Sigma} ,\phi _{l},C_{t})
}\quad \nonumber \\
&=& \cfrac{{\cal N}(x_{t} \mid \mu _{i_{t}}, \Sigma_{i_{t}}){\rm Mult}(i_{t} \mid \phi_{C_{t}})}{\sum_{i_{t}=j} {\cal N}(x_{t} \mid \mu _{j}, \Sigma_{j}){\rm Mult}(j \mid \phi_{C_{t}})}~.
\label{eq:it}
\end{eqnarray}
The prior distribution configured by using the stick breaking process (SBP) \cite{sethuraman1994constructive} is denoted as ${\rm GEM}(\cdot )$, 
the multinomial distribution as ${\rm Mult}(\cdot )$, 
the Dirichlet distribution as ${\rm Dir}(\cdot )$, 
the inverse--Wishart distribution as ${\cal IW}(\cdot )$, and 
the multivariate Gaussian (normal) distribution as ${\cal N}(\cdot )$.
The motion model and the sensor model of self-localization are denoted as $p(x_{t} \mid x_{t-1},u_{t})$ and $p(z_{t} \mid x_{t})$ in equations (\ref{eq:seisei9}) and (\ref{eq:seisei10}), respectively.

This model can learn an appropriate number of spatial concepts, depending on the data, by using a nonparametric Bayesian approach.
We use the SBP, which is one of the methods based on the Dirichlet process.
In particular, this model can consider a theoretically infinite number of spatial concepts $L\rightarrow \infty $ and position distributions $K\rightarrow \infty $.
SBP computations are difficult because they generate an infinite number of parameters.
In this study, we approximate a number of parameters by setting sufficiently large values, i.e., a weak-limit approximation \cite{fox2011sticky}.

It is possible to correlate a name with multiple places, e.g., {\it ``staircase''} is in two different places, and a place with multiple names, e.g., {\it ``toilet''} and {\it ``restroom''} refer to the same place.
Spatial concepts are represented by a word distribution of the names of the place $W_{l}$ and several position distributions ($\mu_{k}$, $\Sigma _{k}$) indicated by a multinomial distribution $\phi _{l}$.
In other words, this model is capable of relating the mixture of Gaussian distributions to a multinomial distribution of the names of places.
{It should be noted that the arrows connecting $i_{t}$ to the surrounding nodes of the proposed graphical model differ from those of ordinal Gaussian mixture model (GMM). We assume that words obtained by the robot do not change its position, but that the position of the robot affects the distribution of words. Therefore, the proposed generative process assumes that the index of position distribution $i_{t}$, i.e., the category of the place, is generated from the position of the robot $x_{t}$. This change can be naturally introduced without any troubles by introducing equation (\ref{eq:it}).}

\begin{table}[tb]
\renewcommand{\arraystretch}{1.2}
\begin{center}
\caption[Each element of the graphical model]{Each element of the graphical model}
\small
\begin{tabular}{|c|c|} \hline
$x_t$ & Self-position of a robot \\ \hline
$u_t$ & Control data \\ \hline
$z_t$ & Sensor data \\ \hline
$C_t$ & Index of spatial concepts \\ \hline
$O_{t,b}$ & Segmented word in a uttered sentence \\ \hline
\raisebox{1ex}{\shortstack{$\mathbf{W}$}} & \raisebox{-0.2ex}{\shortstack{\strut{} Multinomial distribution as the word probability \\of the names of places}} \\ \hline
\raisebox{1ex}{\shortstack{$\mbox{\boldmath $\mu $}, \mathbf{\Sigma}$}} & \raisebox{-0.2ex}{\shortstack{\strut{} Gaussian distribution as a position distribution\\(mean vector, covariance matrix)}} \\ \hline
$i_{t}$ & Index of a position distribution\\ \hline
\raisebox{1ex}{\shortstack{$\phi_{l}$}} & \raisebox{-0.2ex}{\shortstack{\strut{}Multinomial distribution \\of index $i_{t}$ of Gaussian distribution}} \\ \hline
\raisebox{1ex}{\shortstack{$\pi $}} & \raisebox{-0.2ex}{\shortstack{\strut{}Multinomial distribution \\of index $C_t$ of spatial concepts}} \\ \hline
$\alpha $ & Hyperparameter of multinomial distributions $\phi_{l}$ \\ \hline
$\gamma  $ & Hyperparameter of multinomial distribution $\pi $ \\ \hline
$\beta _{0}$ & Hyperparameter of Dirichlet prior distribution \\ \hline
\raisebox{-0.2ex}{\shortstack{$m_{0},\kappa _{0},$ \\ $V_{0},\nu _{0}$}} & \raisebox{-0.4ex}{\shortstack{\strut{}Hyperparameters of \\Gaussian--inverse--Wishart prior distribution}
} \\ \hline 
\end{tabular}
\label{youso}
\end{center}
\end{table}

\subsection{Overview of the proposed method SpCoA}
\label{sec:model:task}
We assume that a robot performs self-localization by using control data and sensor data at all times.
The procedure for the learning of spatial concepts is as follows:
\begin{enumerate}
 \item
An utterer teaches a robot the names of places, as shown in Fig.~\ref{fig:teian} (b).
Every time the robot arrives at a place that was a designated learning target, the utterer says a sentence, including the name of the current place.
 \item
The robot performs speech recognition from the uttered speech signal data. 
Thus, the speech recognition system includes a word dictionary of only Japanese syllables.
The speech recognition results are obtained in a lattice format.
 \item
Word segmentation is performed by using the lattices of the speech recognition results.
 \item
The robot learns spatial concepts from words obtained by word segmentation and robot positions obtained by self-localization for all teaching times.
The details of the learning are given in \ref{sec:learn}.
\end{enumerate}
The procedure for self-localization utilizing spatial concepts is as follows:
\begin{enumerate}
 \item
The words of the learned spatial concepts are registered to the word dictionary of the speech recognition system.
 \item
When a robot obtains a speech signal, speech recognition is performed.
Then, a word sequence as the 1-best speech recognition result is obtained.
 \item
The robot modifies the self-localization from words obtained by speech recognition and the position likelihood obtained by spatial concepts.
The details of self-localization are provided in \ref{sec:mcllcm}.
\end{enumerate}

The proposed method can learn words related to places from the utterances of sentences.
We use an unsupervised word segmentation method latticelm that can directly segment words from the lattices of the speech recognition results of the uttered sentences \cite{neubig2012bayesian}.
The {\it lattice} can represent to a compact the set of more promising hypotheses of a speech recognition result, such as N-best, in a directed graph format.
Unsupervised word segmentation using the lattices of syllable recognition is expected to be able to reduce the variability and errors in phonemes as compared to NPYLM \cite{mochihashi2009bayesian}, i.e., word segmentation using the 1-best speech recognition results.

The self-localization method adopts MCL \cite{dellaert1999monte}, a method that is generally used as the localization of mobile robots for simultaneous localization and mapping (SLAM) \cite{thrun2005probabilistic}.
We assume that a robot generates an environment map by using MCL-based SLAM such as FastSLAM \cite{montemerlo2002fastslam,hahnel2003efficient} in advance, and then, performs localization by using the generated map.
Then, the environment map of both an occupancy grid map and a landmark map is acceptable.
%

\subsection{Learning of spatial concept}
\label{sec:learn}
Spatial concepts are learned from multiple teaching data, control data, and sensor data. 
The teaching data are a set of uttered sentences for all teaching times.
Segmented words of an uttered sentence are converted into a bag-of-words (BoW) representation as a vector of the occurrence counts of words $O_{t,\mathbf{B}}$.
The set of the teaching times is denoted as $T_o=\{t_1,t_2, \ldots ,t_N\}$, and the number of teaching data items is denoted as $N$.
The model parameters are denoted as $\Theta =\{ 
\mathbf{W}, \mbox{\boldmath $\mu $}, \mathbf{\Sigma}, \phi_{l}, \pi \}$.
The initial values of the model parameters can be set arbitrarily in accordance with a condition.
Further, the sampling values of the model parameters from the following joint posterior distribution are obtained by performing Gibbs sampling.
\begin{eqnarray}
p(x_{0:T}, i_{T_{o}}, C_{T_{o}}, \Theta \mid O_{T_{o},{\mathbf {B}}}, u_{1:T}, z_{1:T}, {\mathbf h})
\label{eq:gibbs_all}
\end{eqnarray}
where the hyperparameters of the model are denoted as ${\mathbf h}=\{ \alpha, \gamma, \beta _{0}, m_{0}, \kappa _{0}, V_{0}, \nu _{0} \}$.
The algorithm of the learning of spatial concepts is shown in Algorithm \ref{learn_alg}.

The conditional posterior distribution of each element used for performing Gibbs sampling can be expressed as follows:
An index $i_{t}$ of the position distribution is sampled for each data $t\in T_{o}$ from a posterior distribution as follows:
\begin{eqnarray}
\lefteqn{
i_{t} \sim p(i_{t}=k \mid x_{t},\mbox{\boldmath{$\mu $}},\mathbf{\Sigma},\phi _{l},C_{t})
}\quad \nonumber \\
&\propto {\cal N}(x_{t} \mid \mu _{k=i_{t}}, \Sigma_{k=i_{t}}){\rm Mult}(i_{t}=k \mid \phi_{l=C_{t}}).
\label{eq:gibbs_it}
\end{eqnarray}
An index $C_t$ of the spatial concepts is sampled for each data item $t\in T_{o}$ from a posterior distribution as follows:
\begin{eqnarray}
\lefteqn{
C_{t} \sim p(C_{t}=l \mid x_{t},i_{t},O_{t,\mathbf{B}},\mbox{\boldmath{$\mu $}},\mathbf{\Sigma},\phi _{l},\pi )
}\quad \nonumber \\
&\propto {\rm Mult}(O_{t,\mathbf{B}} \mid W_{l=C_{t}}){\rm Mult}(i_{t}=k \mid \phi_{l=C_{t}}) \nonumber \\
&\hspace{10em} {\rm Mult}(C_{t}=l \mid \pi)
\label{eq:gibbs_Ct}
\end{eqnarray}
where $O_{t,\mathbf{B}}$ denotes a vector of the occurrence counts of words in the sentence at time $t$.
A posterior distribution representing word probabilities of the name of place $\mathbf{W}$ is calculated as follows:
\begin{eqnarray}
\begin{split}
\lefteqn{
p(\mathbf{W} \mid C_{T_{o}},O_{T_{o},\mathbf{B}})
}\quad \\
&\propto  \prod_{l \in L} \biggl[ \prod_{\substack{l=C_t\\ t\in T_o}} p(O_{t,\mathbf{B}}\mid W_{l=C_{t}})\biggr] p(W_{l})
\label{eq:gibbs_W}
\end{split}
\end{eqnarray}
where variables with the subscript $ T_{o} $ denote the set of all teaching times.
A word probability of the name of place $W_{l}$ is sampled for each $l \in L$ as follows:
\begin{eqnarray}
W_{l} \sim {\rm Mult}(\mathbf{O}_{l}\mid W_{l}){\rm Dir}(W_{l} \mid \beta_{0})
\propto {\rm Dir}(W_{l} \mid \beta_{n_l})
\label{eq:gibbs_W_l}
\end{eqnarray}
where $\beta_{n_{l}}$ represents the posterior parameter and $\mathbf{O}_{l}$ denotes the BoW representation of all sentences of $C_{t}=l$ in $t\in T_{o}$.
A posterior distribution representing the position distribution $\mbox{\boldmath{$\mu $}},\mathbf{\Sigma}$ is calculated as follows:
\begin{eqnarray}
\begin{split}
\lefteqn{
p(\mbox{\boldmath{$\mu $}},\mathbf{\Sigma} \mid i_{T_{o}},x_{T_{o}},C_{T_o},\phi _{l})
}\quad \\
&\propto  \prod_{k \in K} \biggl[ \prod_{\substack{k=i_t\\ t\in T_{o}}}
 p(x_{t}\mid \mu _{k=i_{t}},\Sigma _{k=i_{t}})\biggr ]
p(\mu_{k},\Sigma_{k}).
 \label{eq:gibbs_myusig}
\end{split}
\end{eqnarray}
A position distribution $\mu _{k}$, $\Sigma_{k}$ is sampled for each $k \in K$ as follows:
\begin{eqnarray}
\lefteqn{
\mu_{k},\Sigma_{k} \sim {\cal N}(\mathbf{x}_{k}\mid \mu _{k},\Sigma _{k}){\cal NIW}(\mu_{k},\Sigma_{k} \mid m_{0},\kappa _{0},V_{0},\nu_{0}) 
}\quad 
\nonumber \\
&\hspace{35pt}\propto {\cal NIW}(\mu_{k},\Sigma_{k} \mid m_{n_{k}},\kappa _{n_{k}},V_{n_{k}},\nu_{n_{k}})\hspace{5pt}
 \label{eq:gibbs_myusig_k}
\end{eqnarray}
where ${\cal NIW}(\cdot )$ denotes the Gaussian--inverse--Wishart distribution; $m_{n_{k}},\kappa _{n_{k}},V_{n_{k}}$, and $\nu_{n_{k}}$ represent the posterior parameters; and $\mathbf{x}_{k}$ indicates the set of the teaching positions of $i_{t}=k$ in $t\in T_{o}$.
A topic probability distribution $\pi$ of spatial concepts is sampled as follows:
\begin{eqnarray}
\pi \sim {\rm Mult}(C_{T_{o}}\mid \pi){\rm Dir}(\pi \mid \gamma )
\propto {\rm Dir}(\pi \mid C_{T_{o}},\gamma ).
\label{eq:gibbs_pi}
\end{eqnarray}
A posterior distribution representing the mixed weights $\phi_{l}$ of the position distributions is calculated as follows:
\begin{eqnarray}
\begin{split}
\lefteqn{
p(\phi_{l} \mid x_{T_{o}},i_{T_{o}},C_{T_{o}} ,\mbox{\boldmath{$\mu $}},\mathbf{\Sigma})
}\quad \\
&\propto  \prod_{l \in L} \biggl[ \prod_{\substack{l=C_t\\ t\in T_{o}}}
 p(i_{t}\mid \phi_{l=C_{t}})\biggr ]
p(\phi_{l}){.}
\label{eq:gibbs_phi}
\end{split}
\end{eqnarray}
A mixed weight $\phi_{l}$ of the position distributions is sampled for each $l \in L$ as follows:
\begin{eqnarray}
\phi_{l} \sim {\rm Mult}(\mathbf{i}_{l}\mid \phi_{l}){\rm Dir}(\phi_{l} \mid \alpha )
\propto {\rm Dir}(\phi_{l} \mid \mathbf{i}_{l},\alpha  )
\label{eq:gibbs_phi_l}
\end{eqnarray}
where $\mathbf{i}_{l}$ denotes a vector counting all the indices of the Gaussian distribution of $C_{t}=l$ in $t\in T_{o}$.

Self-positions $x_{0:T}$ are sampled by using a Monte Carlo fixed-lag smoother \cite{kitagawa2014computational} in the learning phase. 
The smoother can estimate self-position $x_{0:t}$ and not $p(x_{0:t} \mid u_{1:t}, z_{1:t})$, i.e., a sequential estimation from the given data $u_{1:t}, z_{1:t}$ until time $t$, but it can estimate $p(x_{0:t} \mid u_{1:T}, z_{1:T})$, i.e., an estimation from the given data $u_{1:T}, z_{1:T}$ until time $T$ later than $t$ $(t<T)$.
In general, the smoothing method can provide a more accurate estimation than the MCL of online estimation. 
In contrast, if the self-position of a robot $x_{t}$ is sampled like direct assignment sampling for each time $t$, 
the sampling of $x_{t}$ is divided in the case with the teaching time $t\in T_{o}$ and another time $t\notin T_{o}$ as follows:
\begin{eqnarray}
x_{t} \sim \left\{
\begin{array}{l}
p(x_{t} \mid x_{t-1},x_{t+1},u_{t},u_{t+1},z_{t}) \\
\propto p(x_{t+1}\mid x_{t},u_{t+1})p(z_{t}\mid x_{t})p(x_t\mid x_{t-1},u_t) \\
\hspace{170pt} (t \notin T_{o}){,} \\
p(x_{t} \mid x_{t-1},x_{t+1},u_{t},u_{t+1},z_{t},i_{t},\mbox{\boldmath{$\mu $}},\mathbf{\Sigma},\phi_{l},C_{t}) \nonumber \\
\propto p(x_{t+1}\mid x_{t},u_{t+1})p(z_{t}\mid x_{t})p(x_t\mid x_{t-1},u_t)  \\
\hspace{1.0em} p(i_{t} \mid x_{t},\mbox{\boldmath{$\mu $}},\mathbf{\Sigma} ,\phi _{l},C_{t}) \\ %
\hspace{170pt} (t \in T_{o}){.} 
\end{array} \right.
\hspace{-2em}  \\
\label{eq:gibbs_x_tin}
\end{eqnarray}

\begin{algorithm}[tb]  
\caption{Learning of spatial concepts} 
\label{learn_alg}          
\begin{algorithmic}[1]
\State{${\cal L}=\emptyset$, $T_{o}=\emptyset$}
\State{}
\Comment{Localization and speech recognition}
\For{$t=0$ to $T$}
\State{$x_{0:t}\hspace{-1pt}\sim\hspace{-1pt}{\rm Monte\_Carlo\_smoother}(x_{0:t-1},u_{1:t},z_{1:t})$~\cite{kitagawa2014computational}}
\If{the speech signal is observed} 
\State{$lattice_{t}= {\rm speech\_recognition}({speech\_signal})$}
\State{{add $lattice_{t}$ to ${\cal L}$} }
\Comment{Registering the lattice}
\State{{add $t$ to $T_{o}$} }
\Comment{Registering the teaching time}
\EndIf
\EndFor
\State{}
\Comment{Word segmentation using lattices}
\State $O_{T_{o},{\mathbf B}} \sim {\rm latticelm}({\cal L})$ \cite{neubig2012bayesian}
\State{}
\Comment{Gibbs sampling}
\State{Initialize parameters $i_{T_{o}}$, $C_{T_{o}}$, $\Theta =\{ 
\mathbf{W}, \mbox{\boldmath $\mu $}, \mathbf{\Sigma}, \phi_{l}, \pi \}$}
\For{$j=1$ to $iteration\_number$}
\State{$i_{T_{o}} \sim p(i_{T_{o}} \mid x_{T_{o}},\mbox{\boldmath{$\mu $}},\mathbf{\Sigma} ,\phi _{l},C_{T_{o}})$ \hspace{5pt}(\ref{eq:gibbs_it})}
\State{$C_{T_{o}} \sim p(C_{T_{o}} \mid x_{T_{o}},i_{T_{o}},O_{T_{o},\mathbf{B}},\mbox{\boldmath{$\mu $}},\mathbf{\Sigma} ,\phi _{l},\pi )$ \hspace{5pt}(\ref{eq:gibbs_Ct})}
\State{$\mathbf{W} \sim p(\mathbf{W} \mid C_{T_{o}},O_{T_{o},\mathbf{B}})$ \hspace{5pt}(\ref{eq:gibbs_W_l})}
\State{$\mbox{\boldmath $\mu $}, \mathbf{\Sigma} \sim p(\mbox{\boldmath{$\mu $}},\mathbf{\Sigma} \mid i_{T_{o}},x_{T_{o}},C_{T_o},\phi _{l})$ \hspace{5pt}(\ref{eq:gibbs_myusig_k})}
\State{$\pi \sim p(\pi \mid C_{T_{o}})$ \hspace{5pt}(\ref{eq:gibbs_pi})}
\State{$\phi_{l} \sim p(\phi_{l} \mid x_{T_{o}},i_{T_{o}},C_{T_{o}} ,\mbox{\boldmath{$\mu $}},\mathbf{\Sigma})$ \hspace{5pt}(\ref{eq:gibbs_phi_l})}
\For{$t=0$ to $T$}
\State{$x_{t} \sim \left\{
\begin{array}{l}
p(x_{t} \mid x_{t-1},x_{t+1},u_{t},u_{t+1},z_{t}) \\
\hspace{100pt} (t \notin T_{o}) \\
p(x_{t} \mid x_{t-1},x_{t+1},u_{t},u_{t+1},z_{t}) \\
 p(i_{t} \mid x_{t},\mbox{\boldmath{$\mu $}},\mathbf{\Sigma},\phi_{l},C_{t}) \nonumber \\
\hspace{100pt} (t \in T_{o}) 
\end{array} \right.$ (\ref{eq:gibbs_x_tin})}
\EndFor
\EndFor
\Return $\Theta$
\end{algorithmic}
\end{algorithm}

\subsection{Self-localization of after learning spatial concepts}
\label{sec:mcllcm}
A robot that acquires spatial concepts can leverage spatial concepts to self-localization. 
The estimated model parameters $\Theta =\{ 
\mathbf{W}, \mbox{\boldmath $\mu $}, \mathbf{\Sigma}, \phi_{l}, \pi   \}$ and a speech recognition sentence $O_{t,\mathbf{B}}$ at time $t$ are given to the condition part of the probability formula of MCL as follows: 
\begin{eqnarray}
\lefteqn{
p(x_{0:t}\mid z_{1:t},u_{1:t},O_{1:t,\mathbf{B}},\Theta )
}\quad \nonumber \\
&\propto p(z_t\mid x_{t})p(O_{t,\mathbf{B}}\mid x_t,\Theta ) p(x_t\mid x_{t-1},u_t)\nonumber \\
& \hspace{3em} 
  p(x_{0:t-1}\mid z_{1:t-1},u_{1:t-1},O_{1:t-1,\mathbf{B}},\Theta ) .
\label{eq:mclc}
\end{eqnarray}
When the robot hears the name of a place spoken by the utterer, in addition to the likelihood of the sensor model of MCL, the likelihood of $x_{t}$ with respect to a speech recognition sentence is calculated as follows:
\begin{eqnarray}
\begin{split}
&p(O_{t,\mathbf{B}}\mid x_t,\Theta ) \\
&\propto  \sum_{C_t} \biggl[ p(O_{t,\mathbf{B}}|W_{C_{t}}) \sum_{i_{t}} \Bigl[ p(x_{t}|\mu_{i_{t}}, \Sigma_{i_{t}} )p(i_{t}|\phi _{C_{t}}) \Bigr] p(C_{t}|\pi ) \biggr] .
\nonumber
\label{eq:pox}
\end{split}
\hspace{-2em} \\
\end{eqnarray}

The algorithm of self-localization utilizing spatial concepts is shown in Algorithm \ref{mcl_alg}. 
{The set of particles is denoted as {$X_{t}$}, the temporary set that stores the pairs of the particle {$x_{t}^{[m]}$} and the weight {$w_{t}^{[m]}$}, i.e., $\langle x_{t}^{[m]},w_{t}^{[m]} \rangle$, is denoted as {$\bar{X}_{t}$}.}
The number of particles is $M$.
{The function {${\rm sample\_motion\_model}$} is a function that moves each particle from its previous state {$x_{t-1}$} to its current state {$x_t$} by using control data. The function {${\rm sensor\_model}$} calculates the likelihood of each particle $x_t^{[m]}$ using sensor data~$z_{t}$.
These functions are normally used in MCL. For further details, please refer to \cite{thrun2005probabilistic}.}
In this case, a speech recognition sentence $O_{t,\mathbf{B}}$ is obtained by the speech recognition system using a word dictionary containing all the learned words.
\begin{algorithm}[tb]  
\caption{Self-localization utilizing spatial concepts} 
\label{mcl_alg}                   
\begin{algorithmic}[1]
\Procedure{${\rm Localization}$}{$X_{t-1},u_{t},z_{t},O_{t,\mathbf{B}},\Theta$}
\State $\bar{X}_{t} = X_{t} =  \emptyset$
\For{$m=1$ to $M$}
\State $x_{t}^{[m]} = {\rm sample\_motion\_model}(u_{t},x_{t-1}^{[m]})$ {\hspace{5pt}(\ref{eq:seisei9})}
\State $w_{t}^{[m]} = {\rm sensor\_model}(z_{t},x_{t}^{[m]})$ {\hspace{5pt}(\ref{eq:seisei10})}
\If{the speech signal is observed} 
\State $w_{t}^{[m]} = w_{t}^{[m]} \times   p(O_{t,\mathbf{B}}\mid x_t,\Theta )$
\EndIf
\State {add $\langle x_{t}^{[m]},w_{t}^{[m]} \rangle$ to $\bar{X}_{t}$} 
\EndFor
\For{$m=1$ to $M$}
\State draw $i$ with probability $\propto w_{t}^{[i]}$ 
\State add $x_{t}^{[i]}$ to $X_{t}$
\EndFor
\Return $X_{t}$
\EndProcedure
\end{algorithmic}
\end{algorithm}

\section{Experiment I}
\label{sec:experiments1}
In this experiment, we validate the evidence of the proposed method (SpCoA) in an environment simulated on the simulator platform SIGVerse\footnote{SIGServer-2.2.2, SIGViewer-2.2.0, \url{http://www.sigverse.com/wiki/}} \cite{SIGVerse:SII2010}, which enables the simulation of social interactions.
The speech recognition is performed using the Japanese continuous speech recognition system Julius\footnote{Julius dictation-kit-v4.3.1-linux, GMM-HMM decoding, 
\url{http://julius.sourceforge.jp/}
} \cite{kawahara1998sharable,lee2001julius}. 
The set of 43 Japanese phonemes defined by Acoustical Society of Japan (ASJ)'s speech database committee is adopted by Julius \cite{kawahara1998sharable}.
The representation of these phonemes is also adopted in this study.
The Julius system uses a word dictionary containing 115 Japanese syllables.
The microphone attached on the robot is SHURE's PG27-USB.
Further, an unsupervised morphological analyzer,  
a 
latticelm 0.4, 
 is implemented \cite{neubig2012bayesian}.

In the experiment, we compare the following three types of word segmentation methods.
A set of syllable sequences is given to the graphical model of SpCoA by each method. 
This set is used for the learning of spatial concepts as recognized uttered sentences $O_{T_{o},{\bf B}}$. 

\begin{description}
  \item[(A) latticelm (proposed method)] \mbox{}\\ Syllable recognition results in the lattice format are segmented by using latticelm. 
  \item[(B) 1-best NPYLM] \mbox{}\\ Syllable recognition results of the 1-best method are segmented by using latticelm. In this case, latticelm \cite{neubig2012bayesian} is almost equivalent to NPYLM \cite{mochihashi2009bayesian}. 
  \item[(C) Bag-of-syllables (BoS)] \mbox{}\\ Syllable recognition results of the 1-best method are segmented by each syllable. 
In other words, this method is used for segmenting not words but elements of the recognized syllable sequences directly in the BoS (bag-of-letters) representation.
\end{description}

{The remainder of this section is organized as follows: In Section \ref{sec:sci:gakusyu}, the conditions and results of learning spatial concepts are described. The experiments performed using the learned spatial concepts are described in Section \ref{sec:sci:kiridasi} to \ref{sec:sci:jikoiti}. In Section \ref{sec:sci:kiridasi}, we evaluate the accuracy of the phoneme recognition and word segmentation for uttered sentences. In Section \ref{sec:sci:suiteiseido}, we evaluate the clustering accuracy of the estimation results of index {$C_{t}$} of spatial concepts for each teaching utterance. In Section \ref{sec:sci:basyo}, we evaluate the accuracy of the acquisition of names of places. In Section \ref{sec:sci:jikoiti}, we show that spatial concepts can be utilized for effective self-localization.
}

\subsection{Learning of spatial concepts}
\label{sec:sci:gakusyu}
\subsubsection{Conditions}
\label{sec:sci:gakusyujyouken}
\begin{figure}[tb]
  \begin{center}
    \includegraphics[width=210pt]{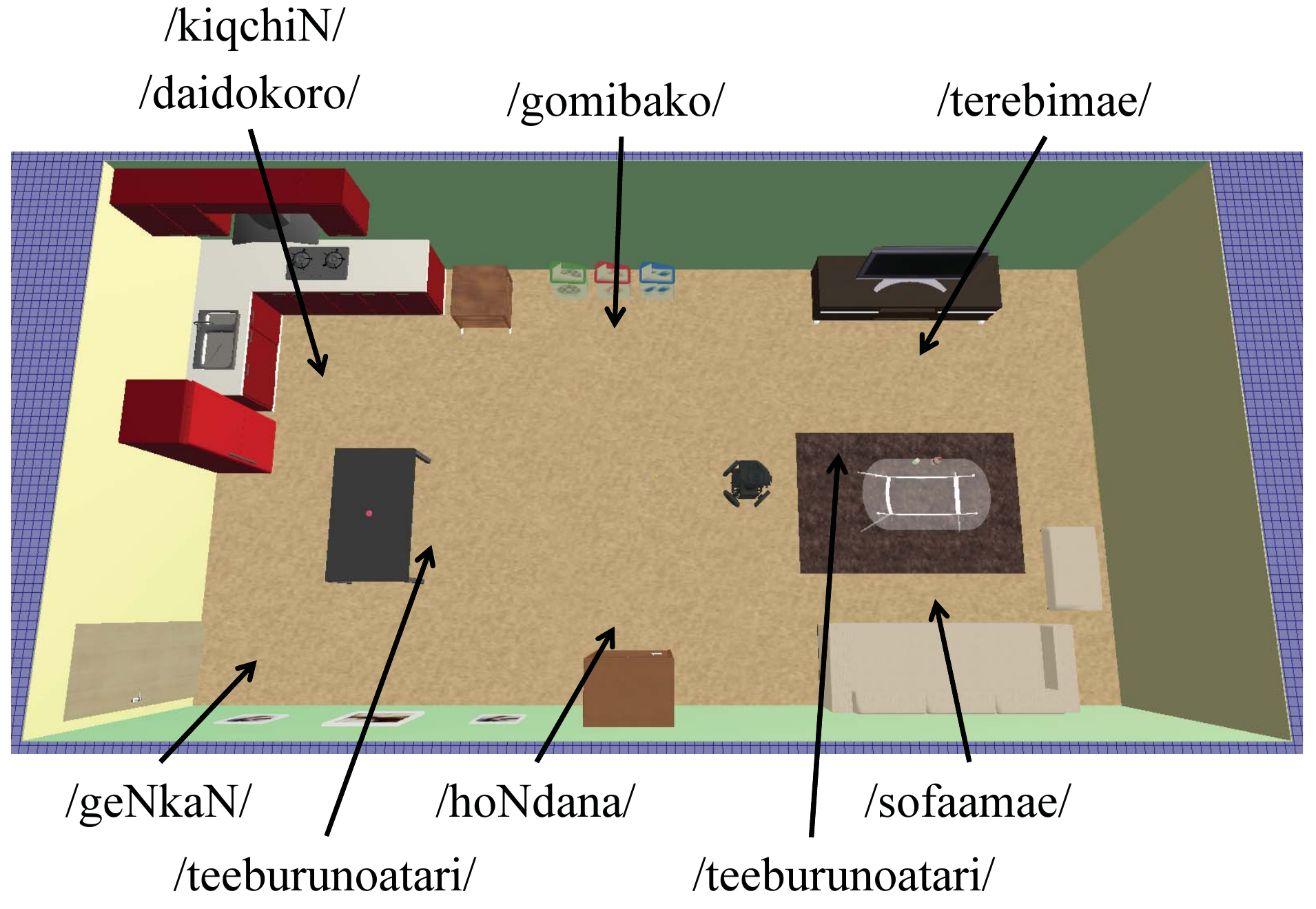}
    \caption{Environment to be used for learning and localization on SIGVerse: This is a pseudo-room in the simulated real world. There is a robot in the center of the room. {
The size of the room is 500~cm $\times$ 1,000~cm, and the size of the robot is 50~cm $\times$ 50~cm.}}
    \label{fig:sekai_sig}
  \end{center}
\end{figure} 
\begin{table}[tb]
\renewcommand{\arraystretch}{1.3}
\begin{center}
\caption[Phrases in each Japanese sentence]{Various phrases of each Japanese sentence: 
{\it ``**''} is used as a placeholder for the name of each place.
Examples of these phrases are {\it ``** is here.''}, {\it ``This place is **.''}, {\it ``This place's name is **.''}, and {\it ``Came to **.''} in English.
}
\small
\begin{tabular}{c|c} \hline \hline
** da{\ }yo &** wa kochira{\ }desu \\ \hline
** desu &kochira{\ }ga ** ni{\ }nari{\ }masu \\ \hline
koko{\ }ga ** &kono{\ }basho{\ }ga ** da{\ }yo \\ \hline
koko{\ }wa ** desu &kono{\ }basho{\ }no{\ }namae{\ }wa ** \\ \hline
** ni{\ }ki{\ }mashi{\ }ta&koko{\ }no{\ }namae{\ }wa ** da{\ }yo \\ \hline \hline
\end{tabular}
\label{iimawashi}
\end{center}
\end{table}
\begin{figure}[!tb]
  \begin{center}
    \includegraphics[width=230pt]{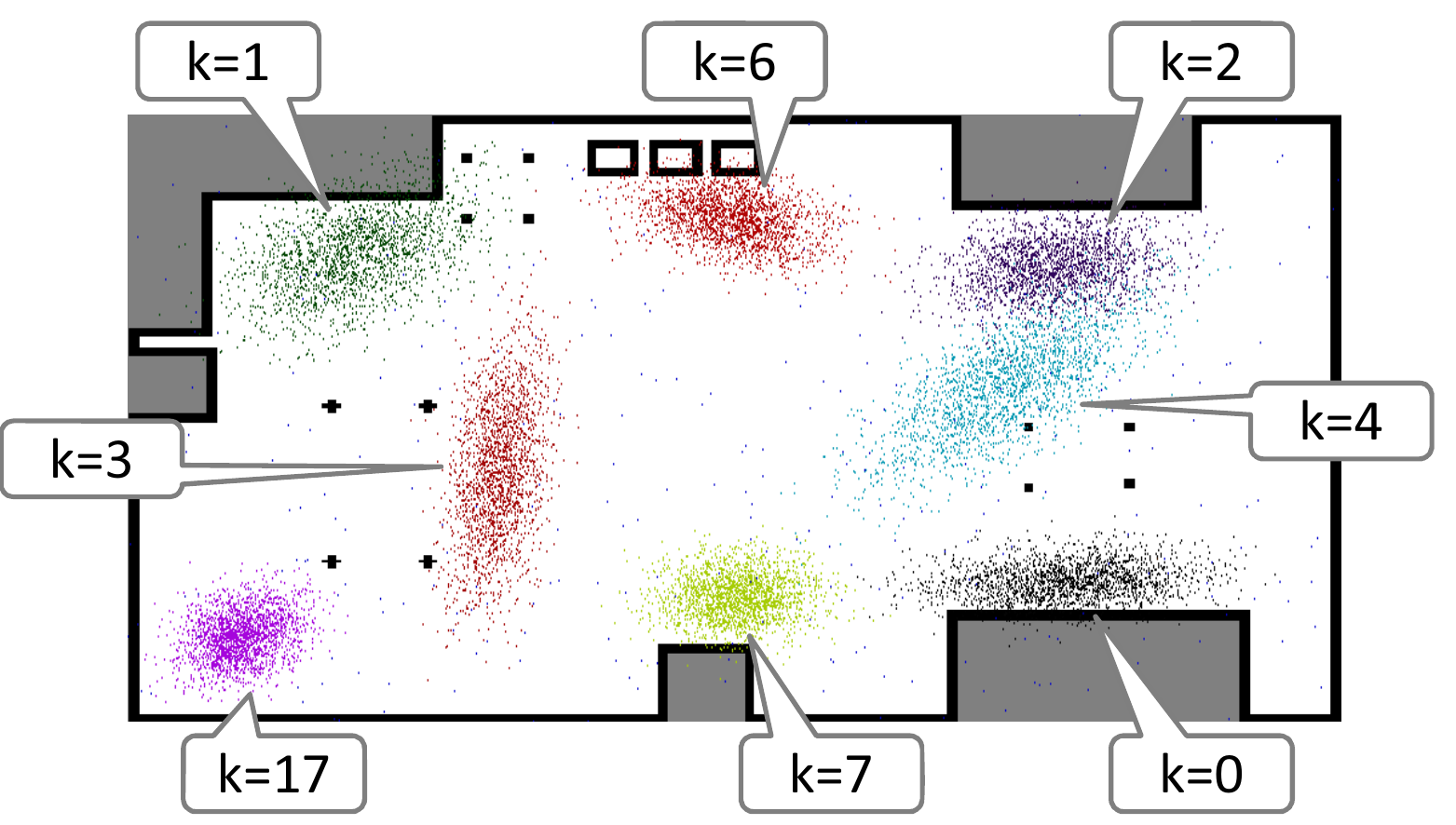}
    \caption{Learning result of the position distribution:   
A point group of each color to represent each position distribution is drawn on an map of the considered environment.
The colors of the point groups are determined randomly.
Each balloon shows the index number for each position distribution.}
    \label{fig:consig}
  \end{center}
  \vspace{15pt}
  \begin{center}
    \includegraphics[width=230pt]{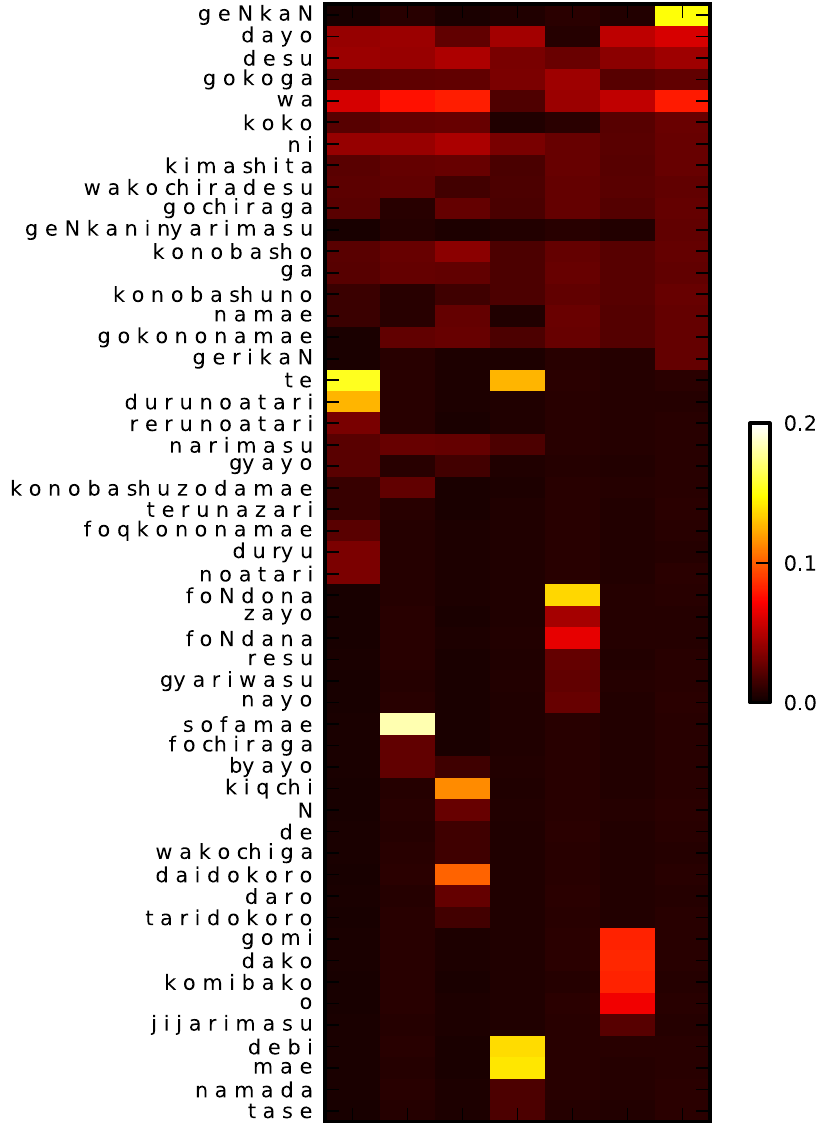}
    \includegraphics[width=230pt]{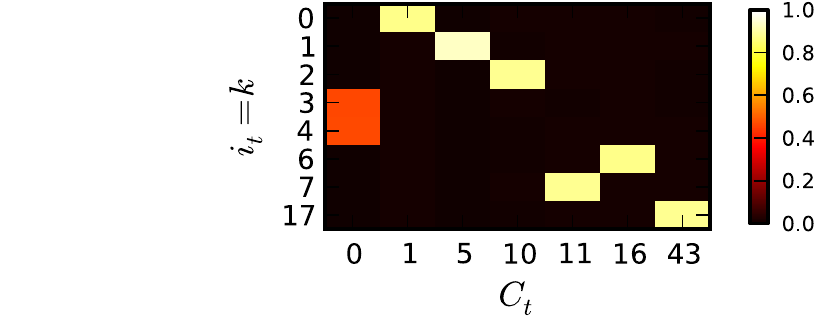}
    \caption{Learning result of the multinomial distributions of the names of places $W$ (top); multinomial distributions of the index of the position distribution $\phi_{l}$ (bottom){: All the words obtained during the experiment are shown.}}
    \label{fig:phi}
  \end{center}
\end{figure}
We conduct this experiment of spatial concept acquisition in the environment prepared on SIGVerse.
The experimental environment is shown in Fig.~\ref{fig:sekai_sig}.
A mobile robot can move by performing forward, backward, right rotation, or left rotation movements on a two-dimensional plane.
In this experiment, the robot can use an approximately correct map of the considered environment.
The robot has a range sensor in front and performs self-localization on the basis of an occupancy grid map.
The initial particles {are defined by the true} initial position of the robot.
The number of particles is $M=1000$.

The lag value of the Monte Carlo fixed-lag smoothing is fixed at 100.
The other parameters of this experiment are as follows:
$L=50$, $K=50$, $\alpha =1.5$, $\gamma =8$, $\beta_{0}=0.5$,  
$m_{0}=[ 0 , 0 ]^{\rm T}
$, $
\kappa_{0}=0.001
$, 
$V_{0}={\rm diag}(1000,1000)$, and 
$\nu_0=2$. 
The number of iterations used for Gibbs sampling is 100. 
This experiment does not include the direct assignment sampling of $x_{t}$ in equation (\ref{eq:gibbs_x_tin}), {i.e., lines 22--24 of Algorithm \ref{learn_alg} are omitted, }because we consider that the self-position can be obtained with sufficiently good accuracy by using the Monte Carlo smoothing.
Eight places are selected as the learning targets, and eight types of place names are considered.
Each uttered place name is shown in Fig.~\ref{fig:sekai_sig}.
These utterances include the same name in different places, i.e., {\it ``teeburunoatari''} (which means {\it near the table} in English), and different names in the same place, i.e., {\it ``kiqchiN''} and {\it ``daidokoro''} (which mean {\it a kitchen} in English).
The other teaching names are {\it ``geNkaN''} (which means {\it an entrance} or {\it a doorway} in English); {\it ``terebimae''} (which means {\it the front of the TV} in English); {\it ``gomibako''} (which means {\it a trash box} in English); {\it ``hoNdana''} (which means {\it a bookshelf} in English); and {\it ``sofaamae''} (which means {\it the front of the sofa} in English).
The teaching utterances, including the 10 types of phrases, are spoken for a total of 90 times.
The phrases in each uttered sentence are listed in Table~\ref{iimawashi}.

\subsubsection{Results}
\label{sec:sci:gakusyukekka}
The learning results of spatial concepts obtained by using the proposed method are presented here.
Fig.~\ref{fig:consig} shows the position distributions learned in the experimental environment.
Fig.~\ref{fig:phi} (top) shows the word distributions of the names of places for each spatial concept, and Fig.~\ref{fig:phi} (bottom) shows the multinomial distributions of the indices of the position distributions.
Consequently, the proposed method can learn the names of places corresponding to each place of the learning target. 
In the spatial concept of index $C_{t}=1$, the highest probability of words was {\it ``sofamae''}, and the highest probability of the indices of the position distribution was $k=0$; therefore, the name of a place {\it ``sofamae''} was learned to correspond to the position distribution of $k=0$.
In the spatial concept of index $C_{t}=5$, {\it ``kiqchi''} and {\it ``daidokoro''} were learned to correspond to the position distribution of $k=1$. Therefore, this result shows that multiple names can be learned for the same place.
In the spatial concept of index $C_{t}=0$, {\it ``te''} and {\it ``durunoatari''} (one word in a normal situation) were learned to correspond to the position distributions of $k=3$ and $k=4$.
Therefore, this result shows that the same name can be learned for multiple places.

\subsection{Phoneme recognition accuracy of uttered sentences}
\label{sec:sci:kiridasi}
\subsubsection{Conditions}
\label{sec:sci:kiridasijyouken}
We compared the performance of three types of word segmentation methods for all the considered uttered sentences.
It was difficult to weigh the ambiguous syllable recognition and the unsupervised word segmentation separately. 
Therefore, this experiment considered the positions of a delimiter as a single letter.
We calculated the matching rate of a phoneme string of a recognition result of each uttered sentence and the correct phoneme string of the teaching data that was suitably segmented {into Japanese morphemes using MeCab\footnote{{MeCab, \url{http://mecab.googlecode.com/svn/trunk/mecab/doc/index.html}}}, which is an off-the-shelf Japanese morphological analyzer that is widely used for natural language processing.}
The matching rate of the phoneme string was calculated by using the phoneme accuracy rate (PAR) as follows:
\begin{eqnarray}
{\rm PAR}= 1 - \frac{S+D+I}{N}.
 \label{eq:ser}
\end{eqnarray}
The numerator of equation (\ref{eq:ser}) is calculated by using the Levenshtein distance between the correct phoneme string and the recognition phoneme string.
$S$ denotes the number of substitutions; $D$, the number of deletions; and $I$, the number of insertions.
$N$ represents the number of phonemes of the correct phoneme string.

\subsubsection{Results}
\label{sec:sci:kiridasikekka}
Table~\ref{table:PAR} shows the results of PAR. 
Table~\ref{bunkatu} presents examples of the word segmentation results of the three considered methods.
We found that the unsupervised morphological analyzer capable of using lattices improved the accuracy of phoneme recognition and word segmentation.
Consequently, this result suggests that this word segmentation method considers the multiple hypothesis of speech recognition as a whole and reduces uncertainty such as variability in recognition by using the syllable recognition results in the lattice format.

\begin{table}[tb]
\renewcommand{\arraystretch}{1.3}
\begin{center}
\caption{Comparison of the phoneme accuracy rates of uttered sentences for different word segmentation methods}
\small
\begin{tabular}{c|c|c|c} \hline
& \shortstack{latticelem\\ (used in SpCoA)} & 1-best NPYLM & BoS\\ \hline
PAR & \underline{{0.82}} & {0.71} & {0.67}  \\ \hline
\end{tabular}
\label{table:PAR}
\end{center}
\end{table}

\begin{table*}[tb]
\renewcommand{\arraystretch}{1.3}
\begin{center}
\caption{Examples of word segmentation results of uttered sentences. {\it ``$\mid$''} denotes a word segment point.}
\small
\begin{tabular}{c|c|c|c} \hline
Correct word sequence & geNkaN$\mid$wa$\mid$kochira$\mid$desu & gomibako$\mid$ni$\mid$ki{$\mid$}mashi{$\mid$}ta & kono{$\mid$}basho$\mid$no$\mid$namae$\mid$wa$\mid$hoNdana\\ \hline
latticelm & geNkaN$\mid$wakochiradesu & komibako$\mid$o$\mid$ni$\mid$kimashita & konobashuno$\mid$namae$\mid$wa$\mid$foNdana\\ \hline
1-best NPYLM & ki$\mid$nika$\mid$N$\mid$wa$\mid$kochira$\mid$de$\mid$su & go$\mid$mibako$\mid$niki$\mid$na$\mid$shita & kono$\mid$bo$\mid$shu$\mid$no$\mid$namae$\mid$wa$\mid$fo$\mid$N$\mid$da$\mid$na\\ \hline
BoS & ki$\mid$ni$\mid$ka$\mid$N$\mid$wa$\mid$ko$\mid$chi$\mid$ra$\mid$de$\mid$su & go$\mid$mi$\mid$ba$\mid$ko$\mid$ni$\mid$ki$\mid$na$\mid$shi$\mid$ta & ko$\mid$no$\mid$bo$\mid$shu$\mid$no$\mid$na$\mid$ma$\mid$e$\mid$wa$\mid$fo$\mid$N$\mid$da$\mid$na\\ \hline
\end{tabular}
\label{bunkatu}
\end{center}
\end{table*}

\subsection{Estimation accuracy of spatial concepts}
\label{sec:sci:suiteiseido}
\subsubsection{Conditions}
\label{sec:sci:suiteijyouken}
We compared the matching rate with the estimation results of index $C_{t}$ of the spatial concepts of each teaching utterance and the classification results of the correct answer given by humans.
The evaluation of this experiment used the adjusted Rand index (ARI) \cite{hubert1985comparing}.
ARI is a measure of the degree of similarity between two clustering results.

Further, we compared the proposed method with a method of word clustering without location information for the investigation of the effect of lexical acquisition using location information.
In particular, a method of word clustering without location information used the Dirichlet process mixture (DPM) of the unigram model of an SBP representation. 
The parameters corresponding to those of the proposed method were the same as the parameters of the proposed method and were estimated using Gibbs sampling.

\subsubsection{Results}
\label{sec:sci:suiteikekka}
\begin{figure}[tb]
  \begin{center}
    \includegraphics[width=250pt]{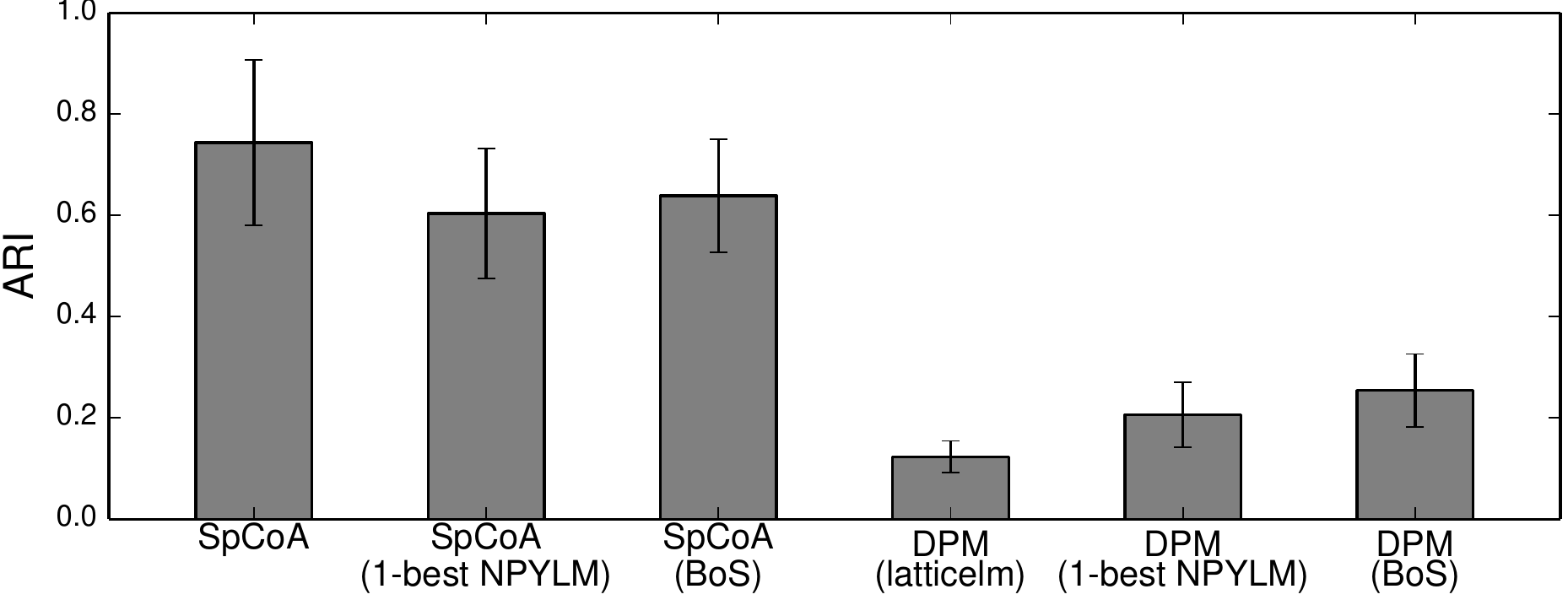}
    \caption{Comparison of the accuracy rates of the estimation results of spatial concepts}
    \label{fig:ARI}
  \end{center}
\end{figure}

Fig.~\ref{fig:ARI} shows the results of the average of the ARI values of 10 trials of learning by Gibbs sampling.
Here, we found that the proposed method showed the best score.
These results and the results reported in Section \ref{sec:sci:kiridasi} suggest that learning by uttered sentences obtained by better phoneme recognition and better word segmentation produces a good result for the acquisition of spatial concepts.
Furthermore, in a comparison of two clustering methods, we found that SpCoA was considerably better than DPM, a word clustering method without location information, irrespective of the word segmentation method used.
The experimental results showed that it is possible to improve the estimation accuracy of spatial concepts and vocabulary by performing word clustering that considered location information.

\subsection{
Accuracy of acquired {phoneme sequences representing} the names of places} 
\label{sec:sci:basyo}
\subsubsection{Conditions}
\label{sec:sci:basyo:jyouken}
We evaluated whether the names of places were properly learned for the considered teaching places.
This experiment assumes a request for the best phoneme sequence $O_{t,{\rm best}}$ representing the self-position $x_{t}$ for a robot.
The robot moves close to each teaching place.
The probability of a word $O_{t,{\rm best}}$ when the self-position $x_{t}$ of the robot is given, $p(O_{t,{\rm best}} \mid x_{t})$, can be obtained by using equation (\ref{eq:pox}).
The word having the best probability was selected.
We compared the PAR with the correct phoneme sequence and a selected name of the place.
Because {\it ``kiqchiN''} and {\it ``daidokoro''} were taught for the same place, the word whose PAR was the higher score was adopted.

\subsubsection{Results}
\label{sec:sci:basyo:kekka}
\begin{figure}[tb]
  \begin{center}
    \includegraphics[width=150pt]{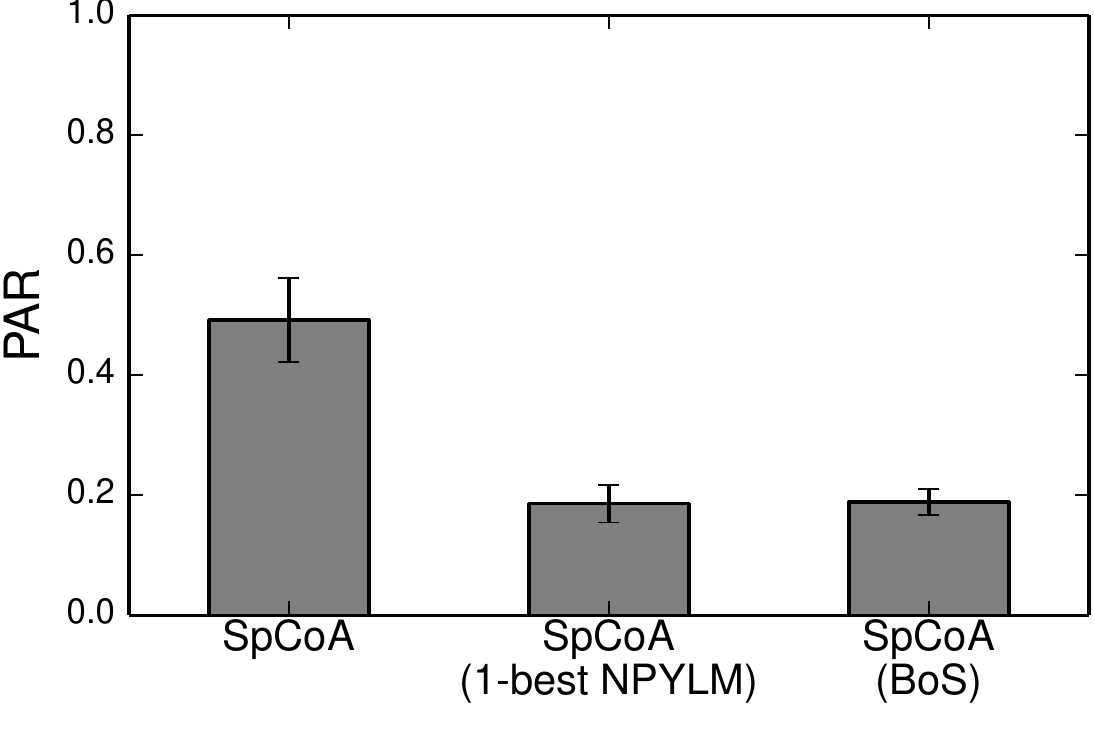}
    \caption{PAR scores for the word considered the name of a place}
    \label{fig:PARw}
  \end{center}
\end{figure}

Fig.~\ref{fig:PARw} shows the results of PAR for the word considered the name of a place. 
SpCoA\,(latticelm), the proposed method using the results of unsupervised word segmentation on the basis of the speech recognition results in the lattice format, showed the best PAR score.
In the 1-best and BoS methods, a part syllable sequence of the name of a place was more minutely segmented as shown in Table~\ref{bunkatu}.
Therefore, the robot could not learn the name of the teaching place as a coherent phoneme sequence.
In contrast, the robot could learn the names of teaching places more accurately by using the proposed method.

\subsection{Self-localization that utilizes acquired spatial concepts}
\label{sec:sci:jikoiti}
\subsubsection{Conditions}
\label{sec:sci:jikoitijyouken}
In this experiment, we validate that the robot can make efficient use of the acquired spatial concepts. 
We compare the estimation accuracy of localization for the proposed method (SpCoA MCL) and the conventional MCL.
When a robot comes to the learning target, the utterer speaks out the sentence containing the name of the place once again for the robot.
The moving trajectory of the robot and the uttered positions are the same in all the trials.
In particular, the uttered sentence is {\it ``kokowa ** dayo''}.
When learning a task, this phrase is not used.
The number of particles is $M=1000$, and the initial particles are uniformly distributed in the considered environment.
The robot performs a control operation for each time step.

The estimation error in the localization is evaluated as follows:
While running localization, we record the estimation error (equation (\ref{eq:et1})) on the ${\bf xy}$ plane of the floor for each time step.
\begin{eqnarray}
e_{t}= \sqrt{(\bar{{\bf x}}_{t} - {\bf x}_{t}^{*})^2 + (\bar{{\bf y}}_{t} - {\bf y}_{t}^{*})^2}
 \label{eq:et1}
\end{eqnarray}
where ${\bf x}_{t}^{*}, {\bf y}_{t}^{*}$ denote the true position coordinates of the robot as obtained from the simulator, and 
$\bar{{\bf x}}_{t}=\sum_{i=1}^{M} w_{t}^{(i)}{\bf x}_{t}^{(i)}$, 
$\bar{{\bf y}}_{t}=\sum_{i=1}^{M} w_{t}^{(i)}{\bf y}_{t}^{(i)}$ represent the weighted mean values of localization coordinates. 
The normalized weight $w_{t}^{(i)}$ is obtained from the sensor model in MCL as a likelihood.
In the utterance time, this likelihood is multiplied by the value calculated using equation (\ref{eq:pox}).
${\bf x}_{t}^{(i)}$, ${\bf y}_{t}^{(i)}$ denote the {${\bf x}$}-coordinate and the {${\bf y}$}-coordinate of index {$i$} of each particle at time {$t$}.
After running the localization, we calculated the average of $e_{t}$.

Further, we compared the estimation accuracy rate (EAR) of the global localization.
In each trial, we calculated the proportion of time step in which the estimation error was less than 50 cm.

\subsubsection{Results}
\label{sec:sci:jikoitikekka}
\begin{figure}[tb]
  \begin{center}
    \includegraphics[width=248pt]{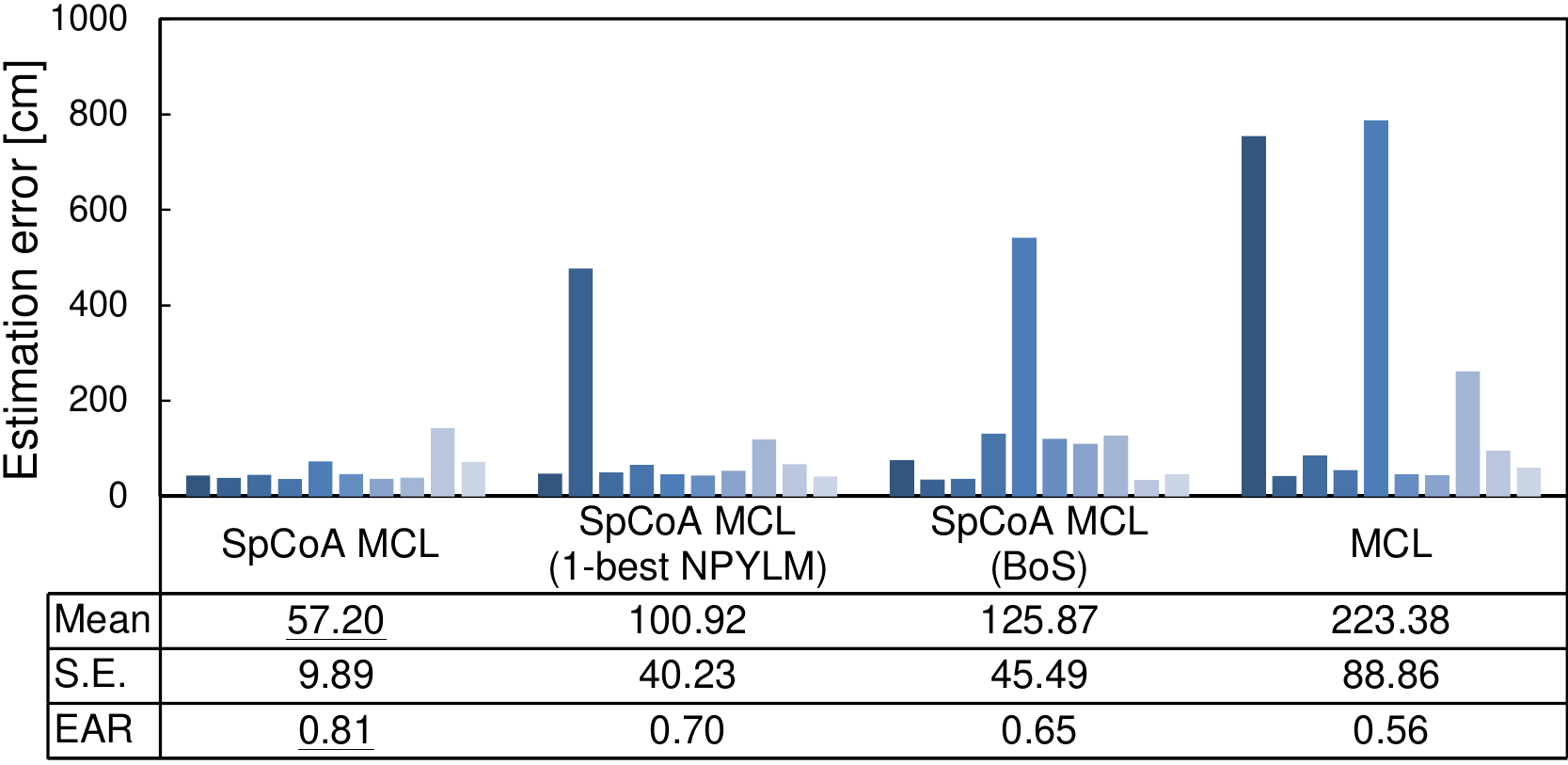}
    \caption{Results of estimation errors and EARs of self-localization}
    \label{fig:mclgt}
  \end{center}
\end{figure}

Fig.~\ref{fig:mclgt} shows the results of the estimation error and the EAR {for 10 trials of each method}. 
All trials of SpCoA MCL\,(latticelm) and almost all trials of the method using 1-best NPYLM and BoS showed relatively small estimation errors. 
{Results of the second trial of 1-best NPYLM and the fifth trial of BoS showed higher estimation errors. In these trials, many particles converged to other places instead of the place where the robot was, based on utterance information.}
Nevertheless, compared with those of the conventional MCL, the results obtained using spatial concepts showed an obvious improvement in the estimation accuracy.
Consequently, spatial concepts acquired by using the proposed method proved to be very helpful in improving the localization accuracy.

\section{Experiment I\hspace{-.1em}I}
\label{sec:experiments3}
\begin{figure}[tb]
  \begin{center}
    \includegraphics[width=100pt]{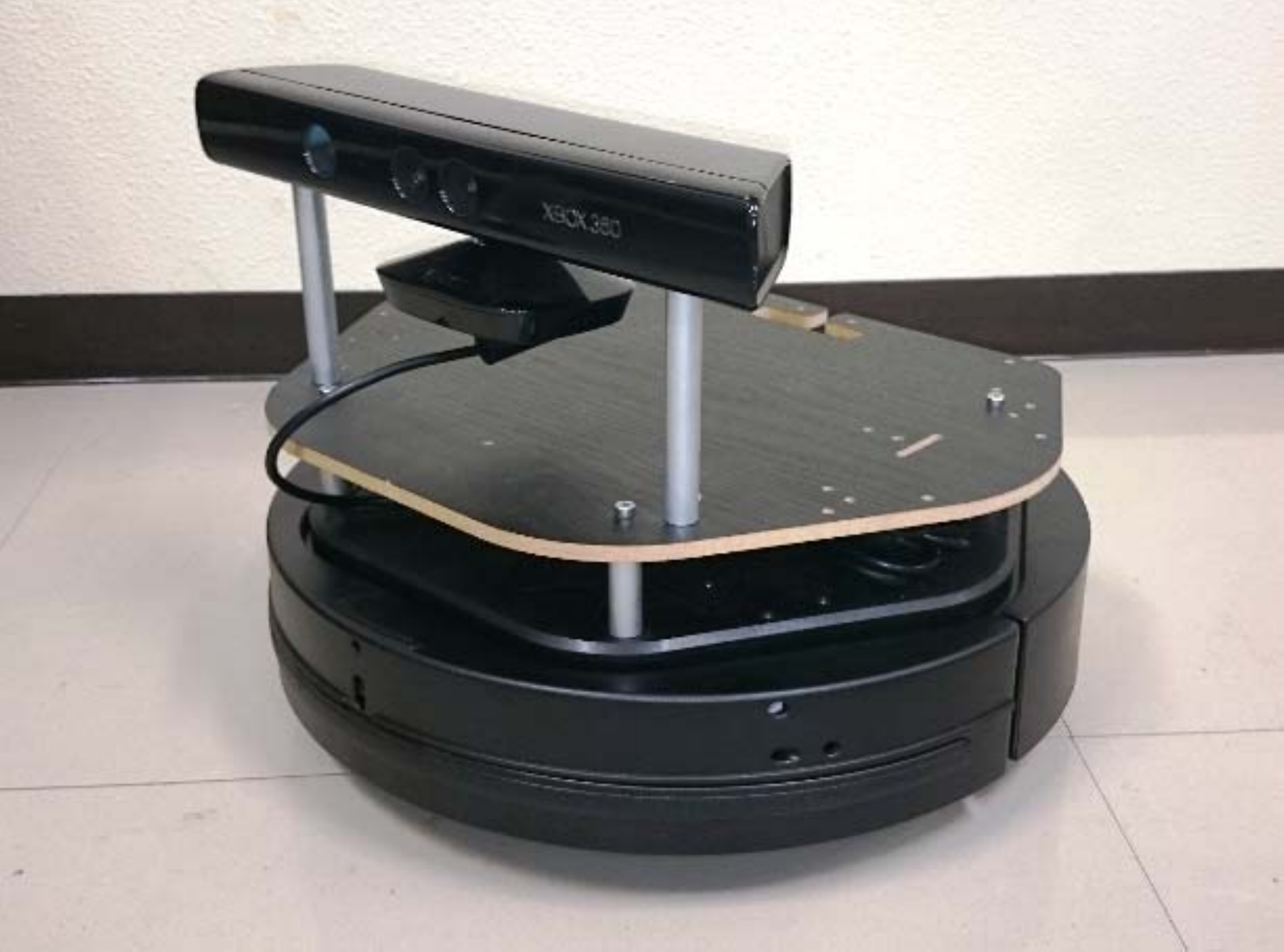}
    \caption{Autonomous mobile robot TurtleBot\,2: The robot is based on Yujin Robot Kobuki and Microsoft Kinect for its use as a range sensor.}
    \label{fig:turtle}
  \end{center}
\end{figure}

In this experiment, the effectiveness of the proposed method was tested by using an autonomous mobile robot TurtleBot\,2\footnote{TurtleBot\,2, \url{http://turtlebot.com/}} in a real environment.
Fig.~\ref{fig:turtle} shows TurtleBot\,2 used in the experiments.
Mapping and self-localization are performed by the robot operating system (ROS).
The speech recognition system, the microphone, and the unsupervised morphological analyzer were the same as those described in Section \ref{sec:experiments1}.

\subsection{Learning of spatial concepts in the real environment}
\label{sec:turtle:gakusyu}
\subsubsection{Conditions}
\label{sec:turtle:gakusyujyouken}
\begin{figure}[tb]
  \begin{center}
    \includegraphics[width=250pt]{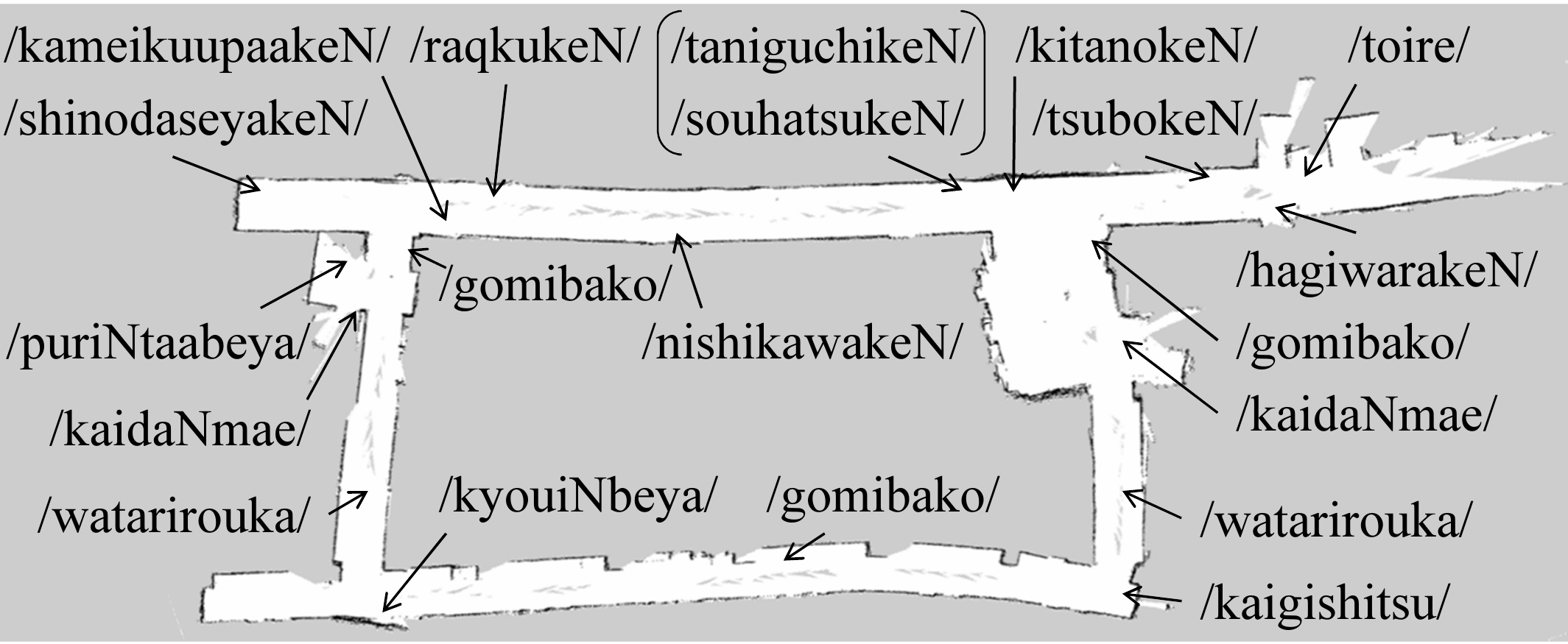}
    \caption{Teaching places and the names of places shown on the generated map.
   The teaching places included places having two names each and multiple places having the same names.}
    \label{fig:map}
  \end{center}
\end{figure}

We conducted an experiment of the spatial concept acquisition in a real environment of an entire floor of a building.
In this experiment, self-localization was performed using a map generated by SLAM.
The initial particles {are defined by the true} initial position of the robot.
The generated map in the real environment and the names of teaching places are shown in Fig.~\ref{fig:map}. 
The number of teaching places was 19, and the number of teaching names was 16.
The teaching utterances were performed for a total of 100 times. 

\subsubsection{Results}
\label{sec:turtle:gakusyukekka}
\begin{figure*}[!tb]
  \begin{center}
    \includegraphics[width=260pt]{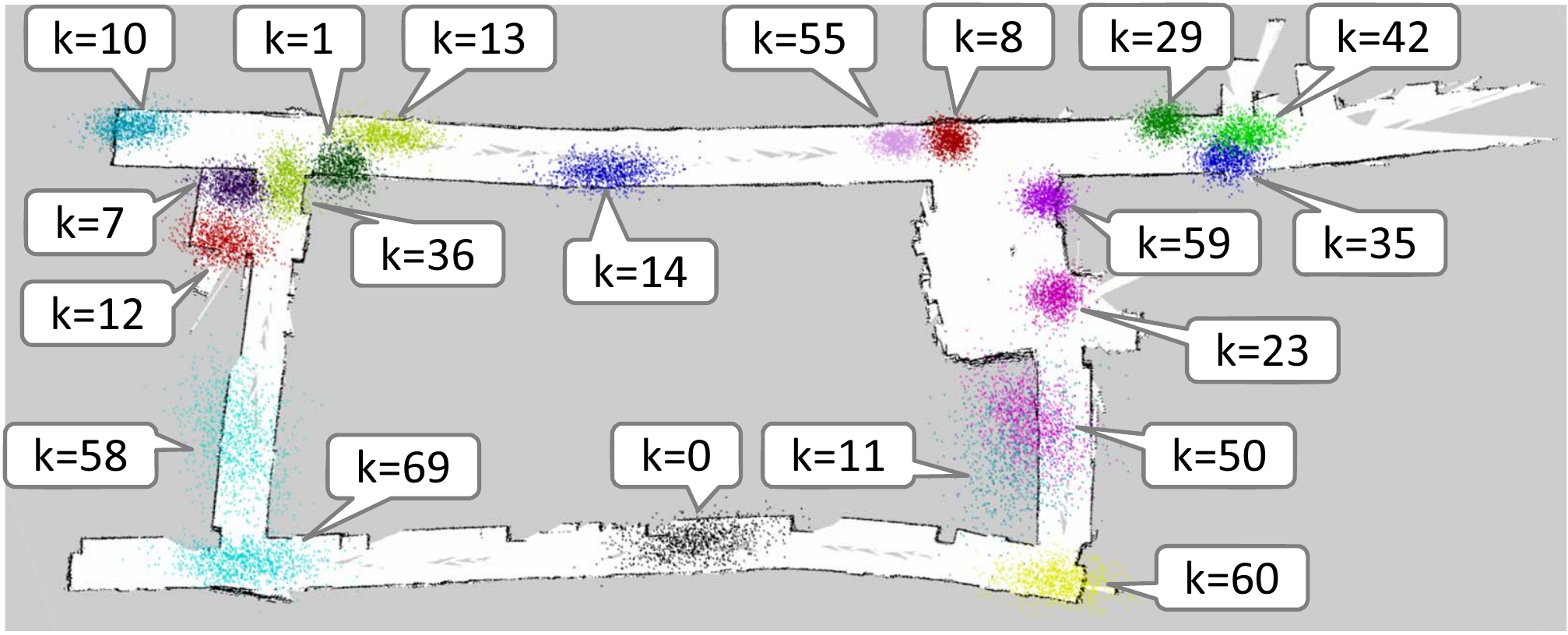}
    \caption{Learning result of each position distribution: A point group of each color denoting each position distribution was drawn on the map.
The colors of the point groups were determined randomly. Further, each index number is denoted as $i_{t}=k$. 
}
    \label{fig:con1n}
  \end{center}
\end{figure*}

Fig.~\ref{fig:con1n} shows the position distributions learned on the map.
Table~\ref{turtle_1} shows the five best elements of the multinomial distributions of the name of place $W_{C_{t}}$ and the multinomial distributions of the indices of the position distribution $\phi _{C_{t}}$ for each index of spatial concept $C_{t}$.
\begin{table*}[!tb]
\begin{center}
\caption{Learning result of high-probability words and indices of the position distribution for each spatial concept}
\begin{tabular}{c}
\begin{minipage}{0.5\hsize}
\begin{center}
\begin{tabular}{|c|cc|cc|} \hline
 & \multicolumn{2}{|c|}{$W_{C_{t}}$} & \multicolumn{2}{|c|}{$\phi_{C_{t}}$} \\ 
Index $C_{t}$ & \multicolumn{2}{|c|}{~~Word  ~~~(Probability)} & Index $i_{t}$ & (Probability) \\ \hline \hline
&  {\bf watarirooka} & ({\bf 0.165}) & {\bf 23} & ({\bf 0.175})  \\  \cline{2-5}
& {\bf kaidaNmae} & ({\bf 0.151}) & {\bf 58} & ({\bf 0.174}) \\  \cline{2-5}
2& desu & (0.117)  & {\bf 50} & ({\bf 0.142}) \\  \cline{2-5}
& nikimashita & (0.069) & {\bf 12} & ({\bf 0.141})  \\  \cline{2-5}
& ewa & (0.068) & {\bf 11} & ({\bf 0.039})  \\ \hline \hline
& {\bf nokeN} & ({\bf 0.189}) & {\bf 29} & ({\bf 0.342})  \\  \cline{2-5}
& {\bf tsu} & ({\bf 0.185}) & 64 & (0.008) \\  \cline{2-5}
3& a & (0.046) & 49 & (0.008) \\  \cline{2-5}
& nayo & (0.045) & 82 & (0.008) \\  \cline{2-5}
& de & (0.045) & 94 & (0.008) \\ \hline \hline
& {\bf shinozaseya} & ({\bf 0.178}) & {\bf 10} & ({\bf 0.339}) \\ \cline{2-5}
& {\bf keN} & ({\bf 0.174}) & 37 & (0.008) \\ \cline{2-5}
6& desu & (0.075) & 5 & (0.008) \\ \cline{2-5}
& koko & (0.042) & 94 & (0.008) \\ \cline{2-5}
& wa & (0.041) & 29 & (0.008) \\ \hline \hline
&  desu & (0.195) & {\bf 36} & ({\bf 0.174})  \\  \cline{2-5}
& {\bf gomibako} & ({\bf 0.180}) & {\bf 59} & ({\bf 0.136}) \\  \cline{2-5}
8& koko & (0.102)  & {\bf 0} & ({\bf 0.136}) \\  \cline{2-5}
& a & (0.081) & 13 & (0.135)  \\ \cline{2-5}
& rapukeN & (0.043) & 9 & (0.006)  \\ \hline \hline
& {\bf torire} & ({\bf 0.154}) & {\bf 42} & ({\bf 0.340}) \\  \cline{2-5}
& wa & (0.118) & 83 & (0.008)\\ \cline{2-5}
10 & kokoga & (0.080) & 9 & (0.008) \\ \cline{2-5}
& nikimashita & (0.045) & 46 & (0.008)\\ \cline{2-5}
& byayo & (0.043) & 11 & (0.008)  \\ \hline \hline
& {\bf tani} & ({\bf 0.098}) & {\bf 55} & ({\bf 0.507}) \\ \cline{2-5}
& {\bf sohatsuke} & ({\bf 0.098}) & 84 & (0.006) \\ \cline{2-5}
27& {\bf N} & ({\bf 0.098}) & 21 & (0.006) \\  \cline{2-5}
& {\bf guchi} & ({\bf 0.096}) & 42 & (0.006) \\ \cline{2-5}
& desu & (0.078) & 37 & (0.006) \\ \hline \hline
& {\bf kidanokeN} & ({\bf 0.209}) & {\bf 8} & ({\bf 0.336}) \\ \cline{2-5}
& desu & (0.091)  & 60 & (0.009) \\ \cline{2-5}
29 & dayo & (0.090) &70 & (0.008)  \\ \cline{2-5}
& a & (0.050) &17 & (0.008)  \\ \cline{2-5}
& konobashoga & (0.050) &2 & (0.008)  \\ \hline
\end{tabular}
\label{turtle_1}
\end{center}
\end{minipage}
\begin{minipage}{0.5\hsize}
\begin{center}
\begin{tabular}{|c|cc|cc|} \hline
 & \multicolumn{2}{|c|}{$W_{C_{t}}$} & \multicolumn{2}{|c|}{$\phi_{C_{t}}$} \\ 
Index $C_{t}$ & \multicolumn{2}{|c|}{~~Word  ~~~(Probability)} & Index $i_{t}$ & (Probability) \\ \hline \hline
& {\bf kaigihitsu} & ({\bf 0.181}) & {\bf 60} & ({\bf 0.301}) \\ \cline{2-5}
& nikimashita & (0.113)  & 0 & (0.065)  \\ \cline{2-5}
32& gomirako & (0.079)  & 7 & (0.064) \\ \cline{2-5}
& a & (0.078)  & 36 & (0.007)  \\\cline{2-5}
& dayo & (0.076)  & 33 & (0.007)  \\ \hline \hline
&{\bf N} & ({\bf 0.113}) & {\bf 1} & ({\bf 0.197}) \\  \cline{2-5}
& {\bf kameikukache} & ({\bf 0.112}) & 36 & (0.075) \\  \cline{2-5}
34& ewa & (0.109) & 42 & (0.075) \\  \cline{2-5}
& konobashunonama & (0.108) & 29 &(0.009) \\  \cline{2-5}
& ninarimasu& (0.043) & 61 & (0.008) \\ \hline \hline
& {\bf rakeN} & ({\bf 0.159}) & {\bf 35} & ({\bf 0.296}) \\  \cline{2-5}
& {\bf hagiwa} & ({\bf 0.157}) & 30 & (0.009) \\  \cline{2-5}
44& desu & (0.080) & 48 & (0.008) \\  \cline{2-5}
& wakochira & (0.046) & 32 & (0.008) \\  \cline{2-5}
& ewa & (0.045) & 68 & (0.008) \\ \hline \hline
& {\bf buriN} & ({\bf 0.132}) & {\bf 7} & ({\bf 0.321}) \\  \cline{2-5}
& {\bf bea} & ({\bf 0.130}) & 0 & (0.067)\\ \cline{2-5}
47 & {\bf pa} & ({\bf 0.107}) & 22 & (0.008) \\ \cline{2-5}
& ewa & (0.083) & 5 & (0.008)\\ \cline{2-5}
& dayo & (0.078) & 96 & (0.008)  \\ \hline \hline
& {\bf wakeN} & ({\bf 0.133}) & {\bf 14} & ({\bf 0.332}) \\  \cline{2-5}
& {\bf nishikya} & ({\bf 0.132}) & 90 & (0.008) \\  \cline{2-5}
66&desu & (0.103) & 25 & (0.008) \\  \cline{2-5}
&a & (0.071) & 69 & (0.008) \\  \cline{2-5}
& {\bf nishi} & ({\bf 0.040}) & 87 & (0.008)  \\ \hline \hline
& {\bf rapukeN} & ({\bf 0.145}) & {\bf 13} & ({\bf 0.173})  \\  \cline{2-5}
& nikimashita & (0.081) & 56 & (0.011)  \\  \cline{2-5}
74& nayo & (0.080) & 2 & (0.010)  \\  \cline{2-5}
& nokeN & (0.017) & 91 & (0.010)  \\  \cline{2-5}
& wakeN & (0.016) & 58 & (0.010)   \\ \hline \hline
&{\bf bea} & ({\bf 0.153}) & {\bf 69} & ({\bf 0.343}) \\  \cline{2-5}
&{\bf N} & ({\bf 0.151}) & 34 & (0.008) \\  \cline{2-5}
75&{\bf kyoi} & ({\bf 0.149}) & 15 & (0.008) \\  \cline{2-5}
&dayo & (0.064) & 75 & (0.008) \\  \cline{2-5}
&desu & (0.062) & 27 & (0.008) \\ \hline
\end{tabular}
\label{turtle_2}
\end{center}
\end{minipage}
\end{tabular}
\end{center}
\end{table*}
Thus, we found that the proposed method can learn the names of places corresponding to the considered teaching places in the real environment.
For example, in the spatial concept of index $C_{t}=10$, {\it ``torire''} was learned to correspond to a position distribution of $k=42$.
Similarly, {\it ``kidanokeN''} corresponded to $k=8$ in $C_{t}=29$, and
{\it ``kaigihitsu''} was corresponded to $k=60$ in $C_{t}=32$.
In the spatial concept of index $C_{t}=27$, 
a part of the syllable sequences was minutely segmented as {\it ``sohatsuke''}, {\it ``N''}, and {\it ``tani''}, {\it ``guchi''}. 
In this case, the robot was taught two types of names.
These words were learned to correspond to the same position distribution of $k=55$.
In $C_{t}=8$, {\it ``gomibako''} showed a high probability, and it corresponded to three distributions of the position of $k=0, 36, 59$. 
{The position distribution of $k = 13$ had the fourth highest probability in the spatial concept $C_{t} = 8$. Therefore, {\it ``raqkukeN,''} which had the fifth highest probability in the spatial concept $C_{t} = 8$ (and was expected to relate to the spatial concept $C_{t}=74$), can be estimated as the word drawn from spatial concept $C_{t}=8$.
However, in practice, this situation did not cause any severe problems because the spatial concept of the index $C_{t}=74$ had the highest probabilities for the word {\it ``rapukeN''} and the position distribution $k=13$ than $C_{t} = 8$. In the probabilistic model, the relative probability and the integrative information are important. When the robot listened to an utterance related to {\it ``raqkukeN,''} it could make use of the spatial concept of index $C_{t}=74$ for self-localization with a high probability, and appropriately updated its estimated self-location.}
We expected that the spatial concept of index $C_{t}=2$ was learned as two separate spatial concepts.
However, {\it ``watarirooka''} and {\it ``kaidaNmae''} were learned as the same spatial concept.
Therefore, the multinomial distribution $\phi _{2}$ showed a higher probability for the indices of the position distribution corresponding to the teaching places of both {\it ``watarirooka''} and {\it ``kaidaNmae''}.

{The proposed method adopts a nonparametric Bayesian method in which it is possible to form spatial concepts that allow many-to-many correspondences between names and places. In contrast, this can create ambiguity that classifies originally different spatial concepts into one spatial concept as a side effect. There is a possibility that the ambiguity of concepts such as {$C_{t}=2$} will have a negative effect on self-localization, even though the self-localization performance was (overall) clearly increased by employing the proposed method. The solution of this problem will be considered in future work.}

{In terms of the PAR of uttered sentences, the evaluation value from the evaluation method used in Section \ref{sec:sci:kiridasi} is 0.83; this value is comparable to the result in Section \ref{sec:sci:kiridasi}. However, in terms of the PAR of the name of the place, the evaluation value from the evaluation method used in Section \ref{sec:sci:basyo} is 0.35, which is lower than that in Section \ref{sec:sci:basyo}. We consider that the increase in uncertainty in the real environment and the increase in the number of teaching words reduced the performance. We expect that this problem could be improved using further experience related to places, e.g., if the number of utterances per place is increased, and additional sensory information is provided.}

\subsection{Modification of localization by the acquired spatial concepts}
\label{sec:turtle:jikoiti}
\subsubsection{Conditions}
\label{sec:turtle:jikoitijyouken}
In this experiment, we verified the modification results of self-localization by using spatial concepts in global self-localization. 
This experiment used the learning results of spatial concepts presented in Section \ref{sec:turtle:gakusyu}. 
The experimental procedures are shown below.
The initial particles were uniformly distributed on the entire floor.
The robot begins to move from a little distance away to the target place.
When the robot reached the target place, the utterer spoke the sentence containing the name of the place for the robot.
Upon obtaining the speech information, the robot modifies the self-localization on the basis of the acquired spatial concepts.
The number of particles was the same as that mentioned in Section \ref{sec:turtle:gakusyu}.

\subsubsection{Results}
\label{sec:turtle:jikoitikekka}
\begin{figure*}[tb]
 \begin{minipage}{0.333\hsize}
  \begin{center}
    \includegraphics[width=170pt]{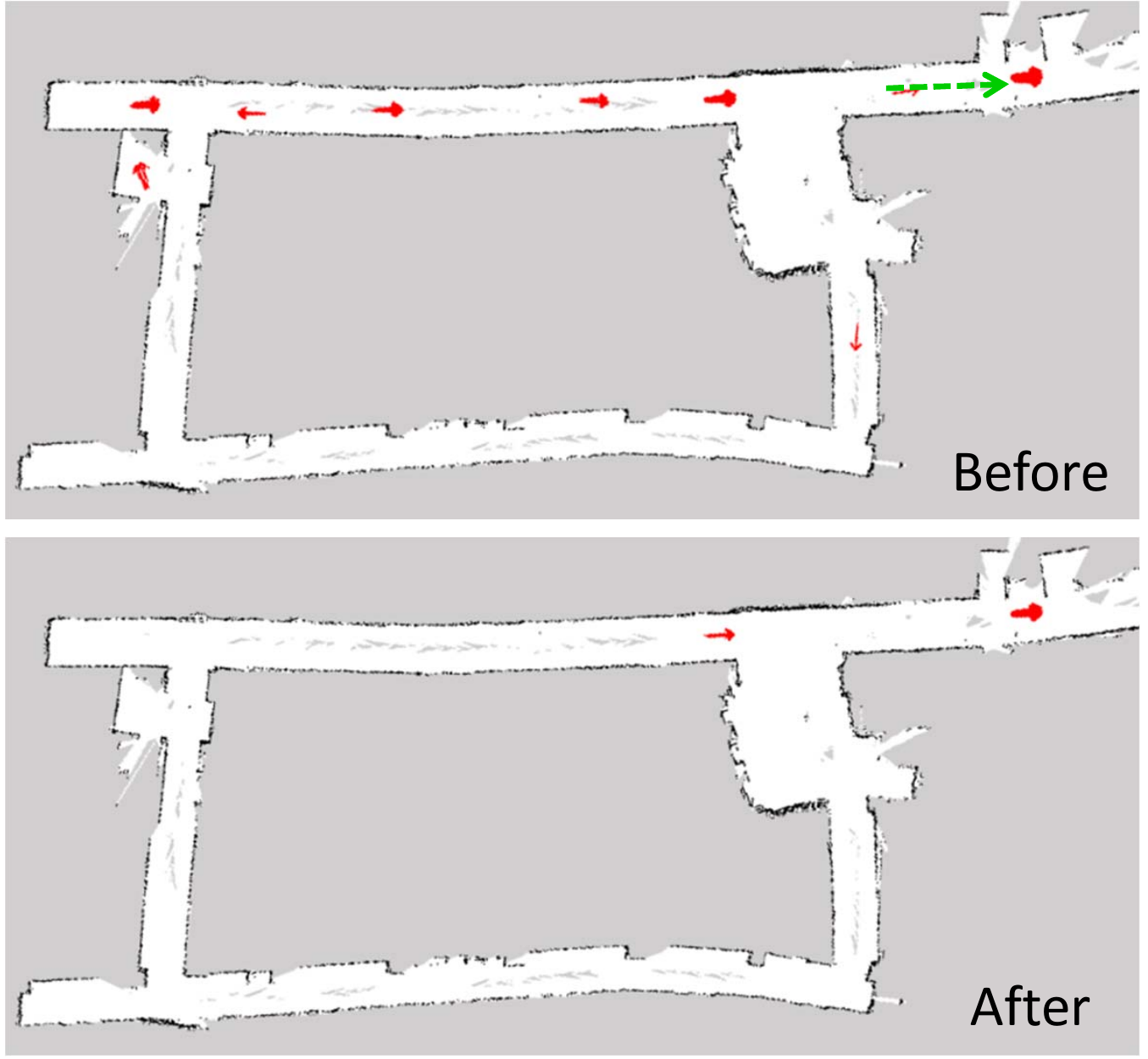}
    \hspace{1.6cm} \small {(a) The name of the place was {\it ``toire.''}}
  \end{center}
   \end{minipage}
 \begin{minipage}{0.333\hsize}
  \begin{center}
    \includegraphics[width=170pt]{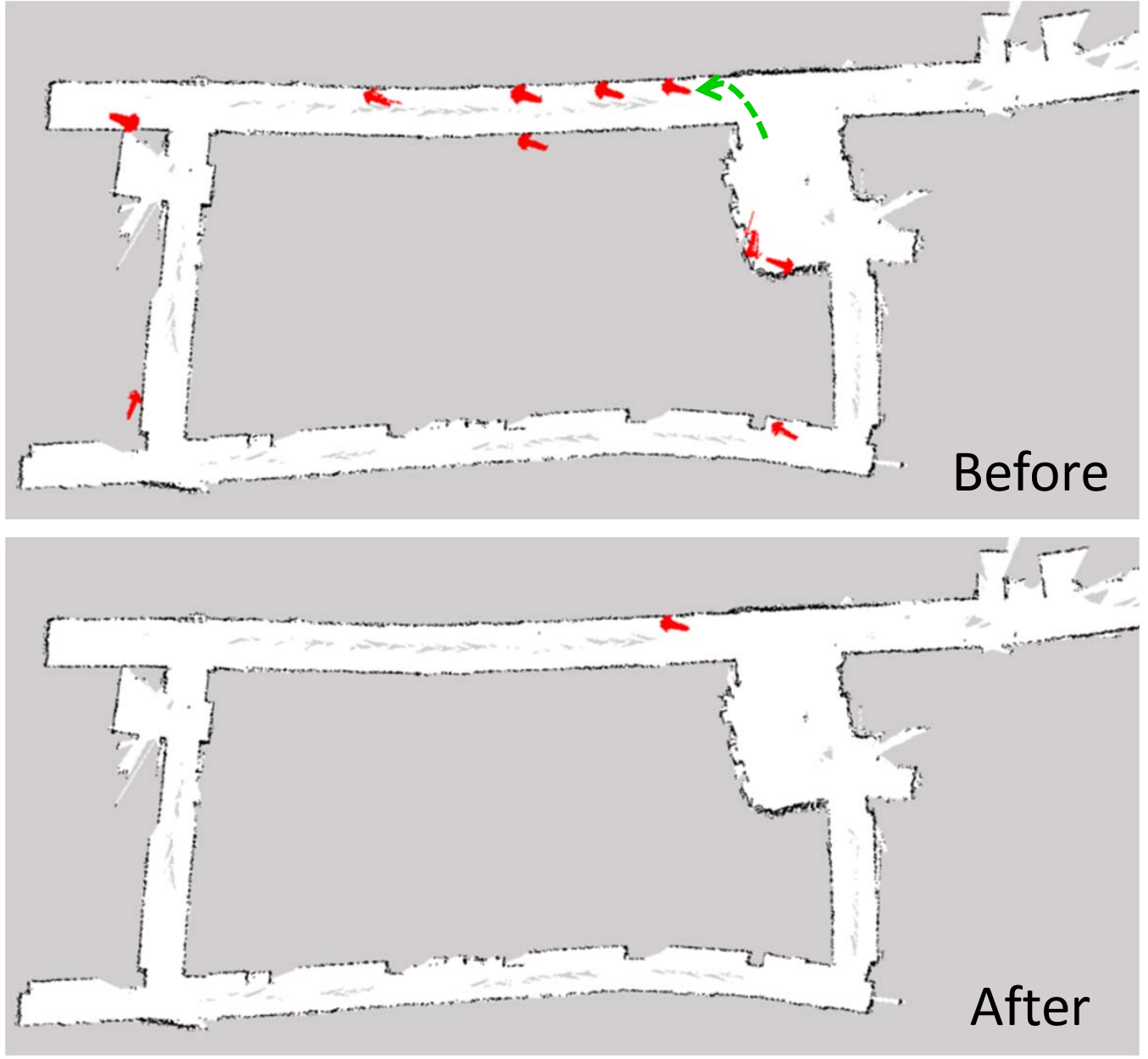}
    \hspace{1.6cm} \small {(b) The name of the place was {\it ``souhatsukeN.''}}
  \end{center}
   \end{minipage}
 \begin{minipage}{0.333\hsize}
  \begin{center}
    \includegraphics[width=170pt]{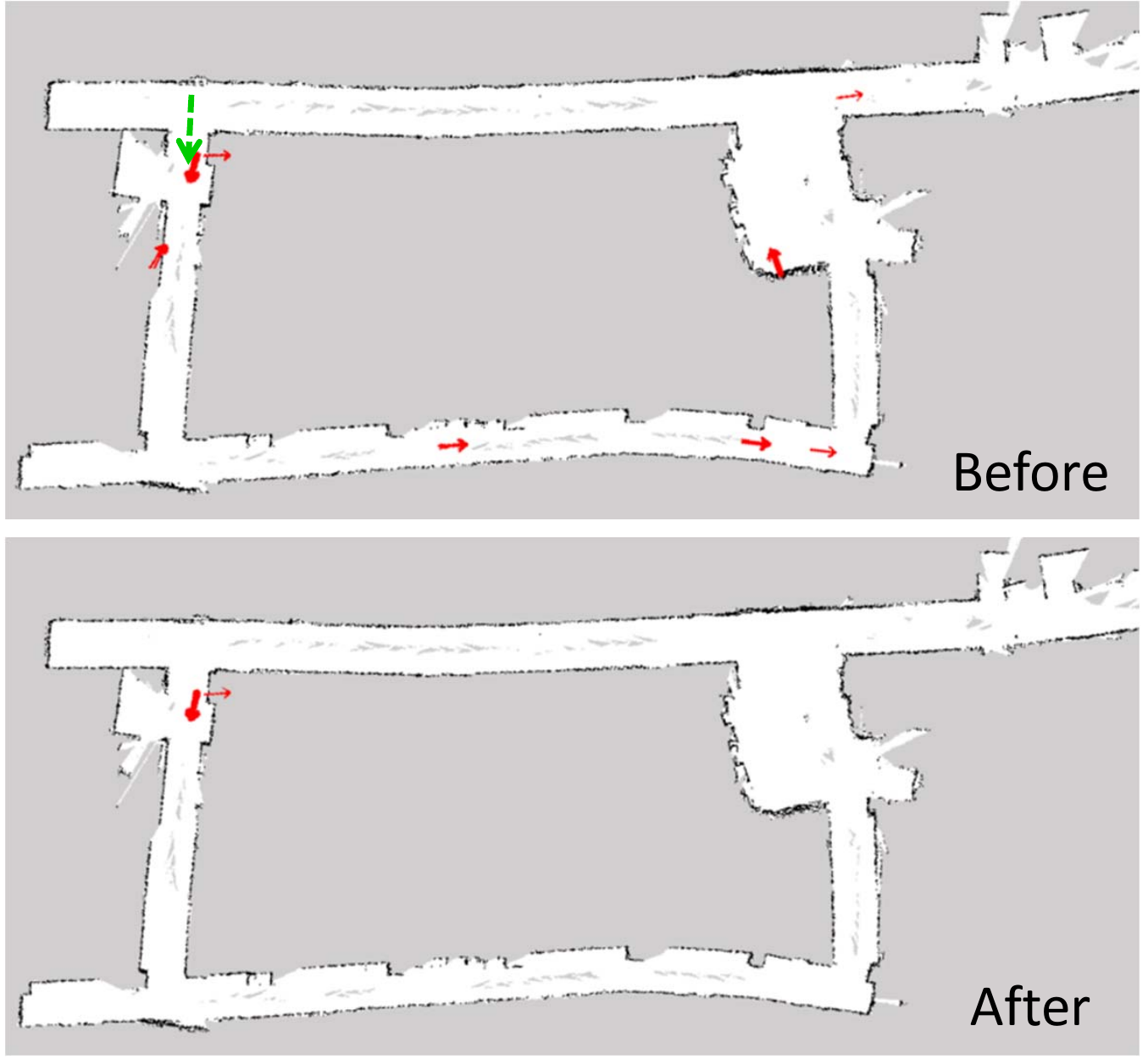}
    \hspace{1.6cm} \small {(c) The name of the place was {\it ``gomibako.''}}
  \end{center}
     \end{minipage}
         \caption{States of particles: before the teaching utterance (top); after the teaching utterance (bottom). The uttered sentence is {\it ``kokowa ** dayo,''} (which means {\it ``Here is **.''}) {\it ``**''} is the name of the place.}
    \label{fig:ji_teisei_all}
\end{figure*}

Fig.~\ref{fig:ji_teisei_all} shows the results of the self-localization before (the top part of the figure) and after (the bottom part of the figure) the utterance for three places.
The particle states are denoted by red arrows.
The moving trajectory of the robot is indicated by a green dotted arrow.
Figs.~\ref{fig:ji_teisei_all} (a), (b), and (c) show the results for the names of places {\it ``toire''}, {\it ``souhatsukeN''}, and {\it ``gomibako''}.
Further, three spatial concepts, i.e., those at $k=0, 36, 59$, were learned as {\it ``gomibako''}.
In this experiment, the utterer uttered to the robot when the robot came close to the place of $k=36$.
In all the examples shown in the top part of the figure, the particles were dispersed in several places.
In contrast, the number of particles near the true position of the robot showed an almost accurate increase in all the examples shown in the bottom part of the figure.
Thus, we can conclude that the proposed method can modify self-localization by using spatial concepts.

\section{Conclusion and Future Work}
\label{sec:conclusion}
In this paper, we discussed the spatial concept acquisition, lexical acquisition related to places, and self-localization using acquired spatial concepts. 
We proposed {\it nonparametric Bayesian spatial concept acquisition method} SpCoA that integrates latticelm\cite{neubig2012bayesian}, a spatial clustering method, and MCL.
We conducted experiments for evaluating the performance of SpCoA in a simulation and a real environment.
SpCoA showed good results in all the experiments.
In experiments of the learning of spatial concepts, the robot could form spatial concepts for the places of the learning targets from human continuous speech signals in both the room of the simulation environment and the entire floor of the real environment. 
Further, the unsupervised word segmentation method latticelm could reduce the variability and errors in the recognition of phonemes in all the utterances.
SpCoA achieved more accurate lexical acquisition by performing word segmentation using the lattices of the speech recognition results.
In the self-localization experiments, the robot could effectively utilize the acquired spatial concepts for recognizing self-position and reducing the estimation errors in self-localization. 

As a method that further improves the performance of the lexical acquisition, 
a mutual learning method was proposed by Nakamura et al. on the basis of the integration of the learning of object concepts with a language model \cite{nakamura2013multimodal,nakamura2014mutual}.
Following a similar approach, Heymann et al. proposed a method that alternately and repeatedly updates phoneme recognition results and the language model by using unsupervised word segmentation \cite{HeWaHa2014}. 
As a result, they achieved robust lexical acquisition.
In our study, we can expect to improve the accuracy of lexical acquisition for spatial concepts by estimating both the spatial concepts and the language model.

Furthermore, as a future work, we consider it necessary for robots to learn spatial concepts online and to recognize whether the uttered word indicates the current place or destination.
{Furthermore, developing a method that simultaneously acquires spatial concepts and builds a map is one of our future objectives. We believe that the spatial concepts will have a positive effect on the mapping.}
We also intend to examine a method that associates the image and the landscape with spatial concepts and a method that estimates both spatial concepts and object concepts.




\bibliographystyle{IEEEtran}   
\bibliography{./TAMD1_a.taniguchi02s_arxiv} 

\end{document}